%% file: tower2024.tex
\documentclass{article} 
\usepackage{tower2024}
\usepackage{CJKutf8}
\usepackage{microtype}
\usepackage{hyperref}
\usepackage{url}
\usepackage{booktabs}
\usepackage{amssymb}
\usepackage{float}

\definecolor{CustomBlue}{RGB}{57,83,191}

\usepackage[most]{tcolorbox}

\newtcbox{\clustertab}[1]{on line, box align=base, colback={#1},colframe={#1},size=fbox,arc=2pt,top=-1.5pt, bottom=-1.5pt, left=-1.5pt, right=-1.5pt, boxrule=0pt, enlarge left by=1pt}

\newcommand{\firstcluster}{{\footnotesize\clustertab{CustomBlue!60}{1}}}
\newcommand{\secondcluster}{{\footnotesize\clustertab{CustomBlue!40}{2}}}
\newcommand{\thirdcluster}{{\footnotesize\clustertab{CustomBlue!25}{3}}}
\newcommand{\fourthcluster}{{\footnotesize\clustertab{CustomBlue!15}{4}}}
\newcommand{\fifthcluster}
{{\footnotesize\clustertab{CustomBlue!10}{5}}}
\newcommand{\sixthcluster}
{{\footnotesize\clustertab{CustomBlue!8}{6}}}
\newcommand{\seventhcluster}
{{\footnotesize\clustertab{CustomBlue!6}{7}}}

\usepackage{arydshln}
\makeatletter
\def\adl@drawiv#1#2#3{%
        \hskip.5\tabcolsep
        \xleaders#3{#2.5\@tempdimb #1{1}#2.5\@tempdimb}%
                #2\z@ plus1fil minus1fil\relax
        \hskip.5\tabcolsep}
\newcommand{\cdashlinelr}[1]{%
  \noalign{\vskip 1.3pt
           \global\let\@dashdrawstore\adl@draw
           \global\let\adl@draw\adl@drawiv}
  \cdashline{#1}[.4pt/2pt]
  \noalign{\global\let\adl@draw\@dashdrawstore
           \vskip 3pt}}
\makeatother

\usepackage{enumitem}
\setlist[itemize,enumerate]{leftmargin=*}

\usepackage{graphicx}
\usepackage{pgf-pie}

\usepackage{pgfplots}
\usetikzlibrary{pgfplots.groupplots}
\pgfplotsset{compat=1.3}
\usepackage{tikz}
\usetikzlibrary{patterns}
\usetikzlibrary{shapes.geometric} 
\usepackage{caption}
\usepackage{subcaption}

\usepackage{colortbl}
\usepackage{xcolor}

\usepackage[most]{tcolorbox}

\title{\textsc{Tower}: An Open Multilingual Large Language\\Model for Translation-Related Tasks}

\author[2,4]{Duarte M. Alves$^\dagger$}
\author[1]{José Pombal$^\dagger$}
\author[1,2,4,5\hspace{60pt}]{Nuno M. Guerreiro$^\dagger$}
\author[1]{Pedro H. Martins}
\author[1]{João Alves}
\author[1]{Amin Farajian}
\author[2,4\hspace{60pt}]{Ben Peters}
\author[1,3]{Ricardo Rei}
\author[2,4,7]{Patrick Fernandes}
\author[2\hspace{100pt}]{Sweta Agrawal$^\star$}
\author[5,6]{Pierre Colombo}
\author[1]{José G.C. de Souza}
\author[1,2,4]{André F.T. Martins}

\affiliation[1]{Unbabel}
\affiliation[2]{Instituto de Telecomunicações}
\affiliation[3]{INESC-ID}
\affiliation[4]{Instituto Superior Técnico \& Universidade de Lisboa (Lisbon ELLIS Unit)}
\affiliation[5]{MICS, CentraleSupélec, Université Paris-Saclay}
\affiliation[6]{Equall}
\affiliation[7]{Carnegie Mellon University}

\abstract{While general-purpose large language models (LLMs) demonstrate proficiency on multiple tasks within the domain of translation, approaches based on open LLMs are competitive only when specializing on a single task. In this paper, we propose a recipe for tailoring LLMs to multiple tasks present in translation workflows. We perform continued pretraining on a multilingual mixture of monolingual and parallel data, creating \TowerBase{}, followed by finetuning on instructions relevant for translation processes, creating \TowerInstruct{}. Our final model surpasses open alternatives on several tasks relevant to translation workflows and is competitive with general-purpose~closed~LLMs. To facilitate future research, we release the \Tower{} models, our specialization dataset, an evaluation framework for LLMs focusing on the translation ecosystem, and a collection of model generations, including ours, on our benchmark.}

\correspondence{$^\dagger$Equal contribution, ordered alphabetically by the first name.\\$^\star$Work partially developed during an internship at Unbabel.\\ \email{duartemalves@tecnico.ulisboa.pt}, \email{\{jose.pombal, nuno.guerreiro\}@unbabel.com}.}

\colmfinalcopy 
\begin{document}
\thispagestyle{firststyle}

\newcommand{\mcfour}{\textsc{mC4}}
\newcommand{\Tower}{\textsc{Tower}}
\newcommand{\TowerBase}{\textsc{TowerBase}}
\newcommand{\TowerInstruct}{\textsc{TowerInstruct}}
\newcommand{\TowerBlocks}{\textsc{TowerBlocks}}
\newcommand{\TowerEval}{\textsc{TowerEval}}
\newcommand{\llama}{LLaMA-2}
\newcommand{\mixtral}{Mixtral-8x7B-Instruct}
\newcommand{\mixtralBase}{Mixtral-8x7B}
\newcommand{\mistral}{Mistral-7B-Instruct-v0.2}
\newcommand{\qwen}{Qwen1.5}
\newcommand{\miqu}{miqu-1}
\newcommand{\gemma}{Gemma}
\newcommand{\gptthreefive}{GPT-3.5-turbo}
\newcommand{\gptfour}{GPT-4}
\newcommand{\alma}{\textsc{Alma}}
\newcommand{\almar}{\textsc{Alma-R}}
\newcommand{\almapretrained}{\textsc{Alma-Pretrain}}
\newcommand{\nllb}{NLLB}

\newcommand{\comet}{\textsc{Comet-22}}
\newcommand{\xcomet}{\textsc{xComet}}
\newcommand{\cometkiwi}{\textsc{CometKiwi-22}}
\newcommand{\cometkiwixxl}{\textsc{CometKiwi-XXL}}
\newcommand{\chrf}{\textsc{chrF}}
\newcommand{\bleurt}{\textsc{Bleurt}}
\newcommand{\errant}{\textsc{ERRANT}}

\newcommand{\flores}{\textsc{Flores-200}}
\newcommand{\wmt}{WMT23}
\newcommand{\tico}{TICO-19}

\maketitle
\section{Introduction}

Many important tasks within multilingual NLP, such as quality estimation, automatic post-edition, or grammatical error correction, involve analyzing, generating or operating with text in multiple languages, and are relevant to various translation workflows --- we call these \textbf{translation-related tasks}. Recently, general-purpose large language models~(LLMs) challenged the paradigm of \textit{per-task} dedicated systems, achieving state-of-the-art performance on several recent WMT shared tasks~\citep{kocmi2023findingswmt,freitagetal2023metrics,nevesetal2023biomedic}. Unfortunately, strong capabilities for \textit{multiple} translation-related tasks have so far been exhibited by \textit{closed} LLMs only~\citep{hendy2023gptmt,kocmi-federmann-2023-gemba,fernandes-etal-2023-devil,raunak-etal-2023-leveraging}. Perhaps because most \textit{open} LLMs are English-centric, approaches leveraging these models still lag behind, having thus far achieved competitive results only when specializing on a \textit{single} task~\citep{xu2024alma,xu-etal-2023-instructscore,iyer-etal-2023-towards}.

\begin{figure}[b]
\centering
\includegraphics[width=.95\textwidth]{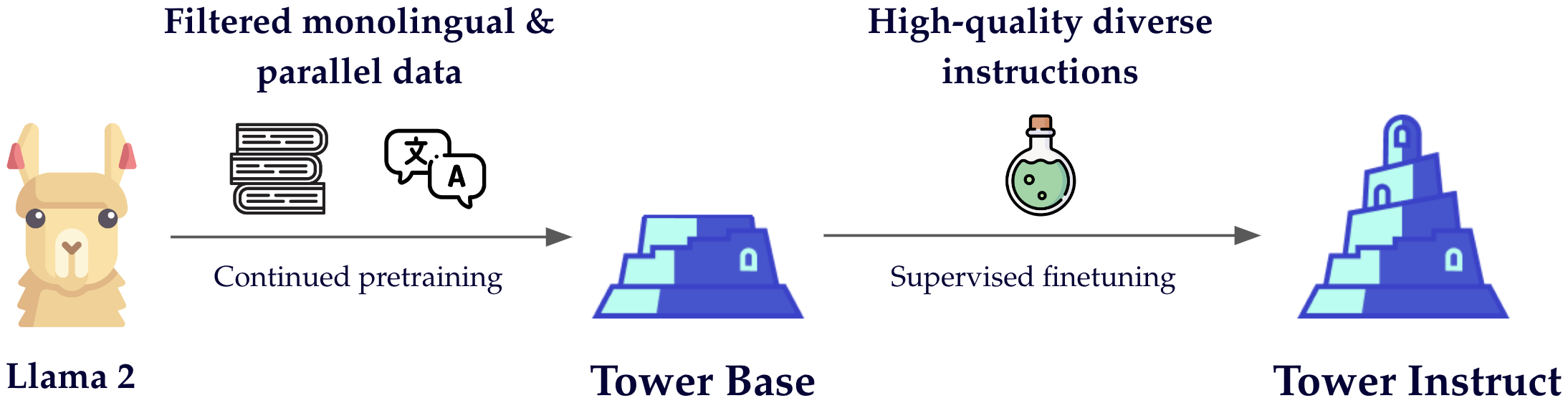}
\caption{Illustration of our method for building \TowerBase{} and \TowerInstruct{}.
}
\label{fig:overview}
\end{figure}

\input{figs/model_comparison_scale}

In this paper, we bridge this gap with a detailed recipe to develop an LLM for \textit{multiple} translation-related tasks. Our approach, illustrated in Figure~\ref{fig:overview} and inspired by \citet{xu2024alma}, relies on three steps. First, we extend the multilingual capabilities of \llama{}~\citep{touvron2023llama2} through continued pretraining on a dataset comprising 20B tokens, creating ~\TowerBase{}~(\S\ref{sec:tower-base}). Importantly, while \citet{xu2024alma} employ a dataset exclusively composed by monolingual data, our approach includes parallel data as an additional cross-lingual signal. Second, we curate a dataset to specialize LLMs for translation-related tasks, \TowerBlocks{}~(\S\ref{sec:tower-blocks}). Third, we perform supervised finetuning to obtain an instruction-following model tailored for the field of translation, \TowerInstruct{}~(\S\ref{sec:tower-instruct}).

We extensively evaluate all our models, comparing with open and closed alternatives on a wide range of tasks~(\S\ref{sec:experiments}).
\TowerInstruct{} consistently achieves higher translation quality than open alternatives and is competitive with the closed GPT-4 and GPT-3.5-turbo models~--- see Figure~\ref{fig:performance-v-scale}. Additionally, \TowerInstruct{} outperforms open models in automatic post-edition, grammatical error correction, and named entity recognition.
Careful ablations also outline the influence of each element in our recipe~(\S\ref{sec:analysis}). We highlight the importance of adding parallel data during continued pretraining for improved translation quality, and the effectiveness of including conversational and coding data on \TowerBlocks{}.

Accompanying this work, we release 1) the \Tower{} family, comprising our \TowerBase{} and \TowerInstruct{} models in the sizes of 7B and 13B; 2) our specialization dataset \TowerBlocks{}; 3) \TowerEval{}, the evaluation framework for LLMs for translation-related tasks that we used to perform all evaluations in this paper; 4) a collection of model of our benchmark to ensure reproducibility and encourage future exploration.\footnote{Links for the \href{https://huggingface.co/collections/Unbabel/tower-659eaedfe36e6dd29eb1805c}{\Tower{} models}; \href{https://huggingface.co/datasets/Unbabel/TowerBlocks-v0.1}{\TowerBlocks{}}; \href{https://github.com/deep-spin/tower-eval}{\TowerEval{}}; Zeno~\citep{cabrera23zeno} \href{https://hub.zenoml.com/project/fd13d5fe-ae80-434c-8bfe-87a80165ea21/Tower MT Generations}{project} with model generations.}

\section{\Tower{}: An Open Multilingual LLM for Translation-Related Tasks}
\label{sec:tower}

Our backbone language model is \llama{}, which is very competitive on a wide range of tasks~\citep{touvron2023llama2} and achieves the best zero-shot translation quality across available open LLMs~\citep{xu2024alma}. Nevertheless, the \llama{} family was exposed to relatively little non-English data during pretraining, limiting its potential for multilingual tasks, such as machine translation.
We alleviate this effect by continuing the pretraining of \llama{} on a highly multilingual corpus~(\S\ref{sec:tower-base}). Afterwards, we introduce our dataset to tailor LLMs for translation-related tasks~(\S\ref{sec:tower-blocks}) and finetune our continued pretrained model to obtain an instruction-following model centered around translation~(\S\ref{sec:tower-instruct}).

\subsection{\TowerBase{}: Extending the multilingual capabilities of \llama{}}
\label{sec:tower-base} 

We extend \llama{}'s training on a highly-multilingual dataset comprising 20 billion tokens --- measured with the model's tokenizer --- for 10 languages: English~(en), German~(de), French~(fr), Dutch~(nl), Italian~(it), Spanish~(es), Portuguese~(pt), Korean~(ko), Russian~(ru), and Chinese~(zh).
While previous work exclusively leverages monolingual data~\citep{xu2024almar}, we draw inspiration from \citet{palm_2, Briakou_Cherry_Foster_2023}, which include parallel data during pretraining. Specifically, we \textit{mix parallel sentences} (one-third) along with monolingual data (two-thirds).
Our results show that this approach greatly benefits translation quality~(\S\ref{sec:analysis}).

\paragraph{Monolingual data.} We collect monolingual data from mC4~\citep{xue2021mT5}, a multilingual web-crawled corpus, uniformly sampling across our languages. Additionally, we \textit{improve data quality} with standard cleaning procedures~\citep{wenzek2019ccnet, touvron2023llama}: deduplication, language identification, and perplexity filtering with KenLM~\citep{heafield-2011-kenlm}.

\paragraph{Parallel Data.} We uniformly sample to-English (xx$\rightarrow$en) and from-English (en$\rightarrow$xx) language pairs from various public sources. Additionally, we \textit{ensure translation quality} by removing sentence pairs below quality thresholds for Bicleaner~\citep{prompsit:2018:WMT,prompsit:2020:EAMT} and \cometkiwi{}~\citep{rei2022cometkiwi} --- we detail parallel data sources and filtering thresholds for monolingual and parallel data in Appendix~\ref{sec:appendix-cp-data}.

\paragraph{Model Training.} We train our models with a codebase based on Megatron-LLM~\citep{epfmgtrn} on 8 A100-80GB GPUs, an effective batch size of 1.57 million tokens per gradient step, and a cosine scheduler with initial and final learning rates of $3\times 10^{-5}$ and $3\times 10^{-6}$, respectively. The training times for \TowerBase{} 7B and 13B were 10 and 20 days.

\subsection{\TowerBlocks{}: A dataset to tailor LLMs for translation-related tasks}
\label{sec:tower-blocks}

\begin{figure}[t]
\centering
\includegraphics[width=.97\textwidth]{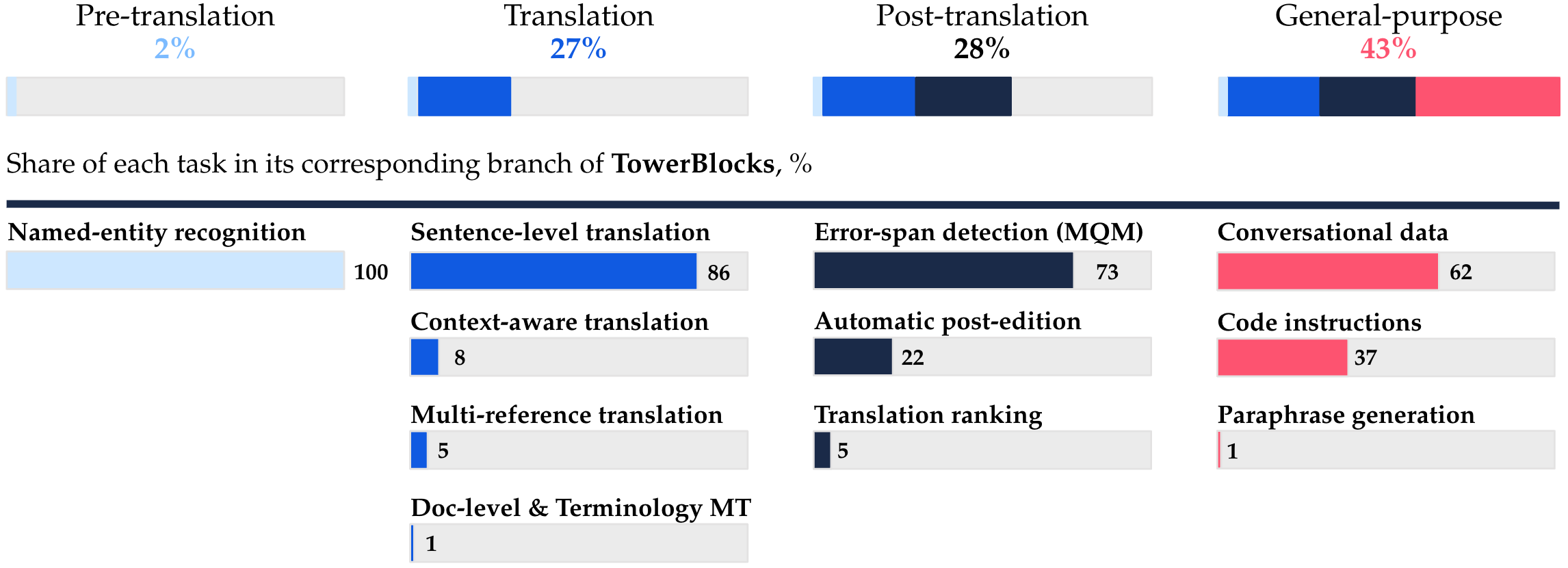}
\caption{Tasks included in our supervised finetuning dataset \TowerBlocks{}.
}
\label{fig:towerblocks-tasks}
\end{figure}

We build \TowerBlocks{} prioritizing data \textit{diversity} and \textit{quality}. Figure~\ref{fig:towerblocks-tasks} illustrates all tasks in the dataset. They include tasks important to translation workflows, applied before or after translation, and datasets to improve multilingual understanding and instruction-following.

\paragraph{Diversity.} We collect records from existing datasets for all translation-related tasks, promoting \textit{domain diversity} by including multiple datasets for each task --- we detail all data sources in Appendix~\ref{sec:tower-blocks-details}. We then reformulate all records as question-answer pairs. Similar to \citet{wei2022flan}, we focus on \textit{template diversity} with multiple manually curated zero- and few-shot templates for each task. Afterwards, we follow the insights from \citet{longpre2022flan}, constructing 75\% of the records as zero-shot instructions. For the remaining records, we include either 1, 3, or 5 in-context examples uniformly sampled from the respective dataset. Finally, we increase \textit{task diversity}, which improves held-in performance up to a moderate number of tasks~\citep{longpre2022flan}, by adding a paraphrasing task, 
dialog data from UltraChat~\citep{ding2023ultrachat}, and coding instructions from Glaive-Code-Assistant.\footnote{\url{https://huggingface.co/datasets/glaiveai/glaive-code-assistant}}

\paragraph{Quality.} Similar to \citet{xu2024alma}, we construct our question-answer pairs from \textit{human-annotated records},\footnote{For named entity recognition, we did not find a permissively licensed human-annotated dataset, so we use MultiCoNER~\citep{malmasi-etal-2022-multiconer,fetahu-etal-2023-multiconer}. For general translation, we include a small amount of parallel data from OPUS to cover all language pairs. Nevertheless, we apply Bicleaner using a threshold of 0.85 followed by the quality filtering procedure described in this section.} prioritizing validation or older test sets. Importantly, we ensure that records from 2023 onwards are excluded from the training data.
We also \textit{avoid reference quality issues}~\citep{xu2024almar} for tasks with reference translations, such as translation and automatic post-edition, by scoring source-reference pairs with \textsc{xComet-QE-Ensemble}~\citep{guerreiro2023xcomet} and discarding records with quality scores below $0.85$.
Additionally, we \textit{avoid  translationese} on the source side, which is associated with numerous quality issues~\citep{zhang-toral-2019-effect,riley-etal-2020-translationese}, by only including translation pairs in their original direction.
Finally, we adopt the UltraChat~\citep{ding2023ultrachat} dialogues filtered by \citet{tunstall2023zephyr} and additionally exclude records respective to translation requests, conversations with formatting issues~(e.g., instructions starting with punctuation, and others), and instances where the assistant refuses to answer.

\subsection{\TowerInstruct{}: Specializing \TowerBase{} for Translation-Related Tasks}
\label{sec:tower-instruct}

As a final step, we obtain \TowerInstruct{} by finetuning \TowerBase{} on \TowerBlocks{}.

\paragraph{Dialog template.} We format each dialog as a single tokenizable string using the \texttt{chatml} template~\citep{chatml}; we provide an example in Appendix~\ref{sec:appendix-chat-template}. This template clearly separates between instructions and answers, and allows for mutli-turn dialog. The template has three special identifiers (control tokens) to delimit messages: \textcolor{CustomBlue}{\texttt{<|im\_start|>user}} and \textcolor{CustomBlue}{\texttt{<|im\_start|>assistant}} preempt the beginning of a turn, and \textcolor{CustomBlue}{\texttt{<|im\_end|>}} marks its end.
We avoid the separation of \textcolor{CustomBlue}{\texttt{<|im\_start|>}} and \textcolor{CustomBlue}{\texttt{<|im\_end|>}} into multiple tokens by extending the tokenizer for \TowerInstruct{} with two dedicated tokens. We do not explicitly add new tokens for \textcolor{CustomBlue}{\texttt{user}} and \textcolor{CustomBlue}{\texttt{assistant}}, as both strings already have dedicated tokens. Additionally, we overwrite the end-of-sequence token with the \textcolor{CustomBlue}{\texttt{<|im\_end|>}} token.

\paragraph{Model training.} We finetune the model with the standard cross-entropy loss, enabling bfloat16 mixed precision and packing~\citep{raffel2020t5}. We only calculate the loss on target (answer) tokens. We train for 4 epochs using a low learning rate and a large batch size --- we detail all hyperparameters in Appendix~\ref{sec:appendix-tower-instruct-hyperparams}. We found that this combination performed the best and eliminated step-wise training losses that have been observed in recent models~\citep{tunstall2023zephyr,kaokao2023intelneuralchat}.\footnote{One hypothesis put forward in~\citet{fastai-2023-can} is that LLMs can rapidly memorize examples during training with one gradient step. In fact, the sudden downward shifts in loss occur precisely when a new epoch starts.} 
Our training took around 50h on 4 NVIDIA A100-80GB GPUs and leveraged the Axolotl framework\footnote{\url{https://github.com/OpenAccess-AI-Collective/axolotl}} and DeepSpeed~\citep{deepspeed} for model parallelism.

\section{Experiments}\label{sec:experiments}

\subsection{Experimental Setup}

\paragraph{Datasets and Tasks.} We analyze translation capabilities using \flores{}~\citep{nllbteam2022}, WMT23~\citep{kocmi2023findingswmt}, and \tico~\citep{anastasopoulos-etal-2020-tico}.
Additionally, we examine three translation-related tasks. First, we evaluate automatic post-edition (APE) by measuring final translation quality after post-editing \nllb{}-3.3B~\citep{nllbteam2022} translations for WMT23.
Second, we evaluate named entity recognition (NER), useful for entity anonymization, using the test split from MultiCoNER 2023~\citep{fetahu-etal-2023-multiconer}.\footnote{We uniformly sample 1000 of the more than 200k records due to the computational costs of evaluating all models on the whole test set.}
Third, we evaluate grammatical error correction (GEC), which is \textit{held out} from our training data and can be applied to correct the source sentence before translation.
We test GEC on CoNLL-2014~\citep{ng2014conll} (English), COWSL2H~\citep{COWSL2H2020} (Spanish), and mlconvgec2018~\citep{chollampatt2018mlconv} (German).

\paragraph{Baselines.} On all tasks, we compare the \Tower{} models with the open models \llama{}~70B~\citep{touvron2023llama2} and \mixtral{}~\citep{jiang2024mixtral}, and the closed-source models \gptthreefive{} and \gptfour{}.\footnote{We use \texttt{gpt-3.5-turbo-0613} and \texttt{gpt-4-0613} available from the official OpenAI API.} For the task of machine translation, we also compare with dedicated systems \nllb{}-54B~\citep{nllbteam2022} and \almar{}~\citep{xu2024almar}. We also report numbers on other open alternatives --- \gemma{}~7B~\citep{gemma-2024}, \mistral{}~\citep{jiang2023mistral} and \qwen{}~72B~\citep{qwen} --- in Appendix~\ref{sec:appendix-translation}.\footnote{\TowerInstruct{} outperforms all these open alternatives.} All model generations are performed with greedy decoding --- we explore alternative decoding methods in Appendix~\ref{sec:appendix-decoding}. For \llama{}~70B and \mixtral{}, we always provide 5 in-context learning examples randomly selected from the development set in the prompt. Unless specified, we evaluate all other models in a 0-shot fashion.

\paragraph{Evaluation.} We evaluate translation quality with \comet{}~\citep{rei-etal-2022-comet} for both MT and APE. For translation, we also report \xcomet{}~\citep{guerreiro2023xcomet}, \cometkiwi{}~\citep{rei2022cometkiwi}, \bleurt{}~\citep{sellam-etal-2020-bleurt}, and \chrf{}~\citep{popovic-2015-chrf} in Appendix~\ref{sec:appendix-translation}.\footnote{We find that performance trends largely hold across metrics. Yet, there is a significant quality gap between \almar{} and \Tower{} models in terms of \chrf{} --- e.g., over 7 points in en$\rightarrow$xx directions on \wmt{}  --- which is not found with neural metrics. We posit that \almar{}’s alignment process on translations preferred by \cometkiwixxl{}~\citep{rei-etal-2023-scaling} and \xcomet{} may inadvertently degrade performance on lexical metrics. Exploring evaluation dynamics after alignment with translation quality metrics is a promising direction for future work.}
For GEC, we measure edit rate (ER)~\citep{snover2006study} and report \errant\citep{errant1,errant2} in Appendix~\ref{sec:appendix-related-tasks}. For NER, we measure sequence F1 score.
On all tasks, we also report performance clusters based on statistically significant performance gaps. For a given language, we verify whether measured differences between all system pairs are statistically different.\footnote{We apply significance testing at a confidence threshold of 95\%. For segment-level metrics such as \comet{} we can perform significance testing at the segment level. However, for corpus-level metrics such as ER and Sequence F1, we follow \citet{koehn-2004-statistical} and perform bootstrapping with 100 samples of size 500 each, applying significance testing on the sample scores.} Afterwards, we create \textit{per-language} groups for systems with similar performance by following the clustering procedure in \citet{freitagetal2023metrics}. Finally, we obtain system-level rankings across multiple languages using a normalized Borda count~\citep{colombo2022best}, which is defined as an average of the obtained clusters.
Note that a first cluster will not exist if no model significantly outperforms all others on a majority of languages.

\subsection{Translation}\label{sec:experiments-mt}

\begin{table}[t]
\begin{center}
\include{tables/translation_results}
\end{center}
\caption{Results for machine translation aggregated by language pair. Models with statistically significant performance improvements are grouped in quality clusters. We highlight the best ranked models in bold and underline the best ranked open models.}
\label{tab:wmt-tico-results}
\end{table}

\begin{figure*}[t]
\begin{subfigure}[t]{0.495\textwidth}
\centering
\includegraphics[width=\linewidth]{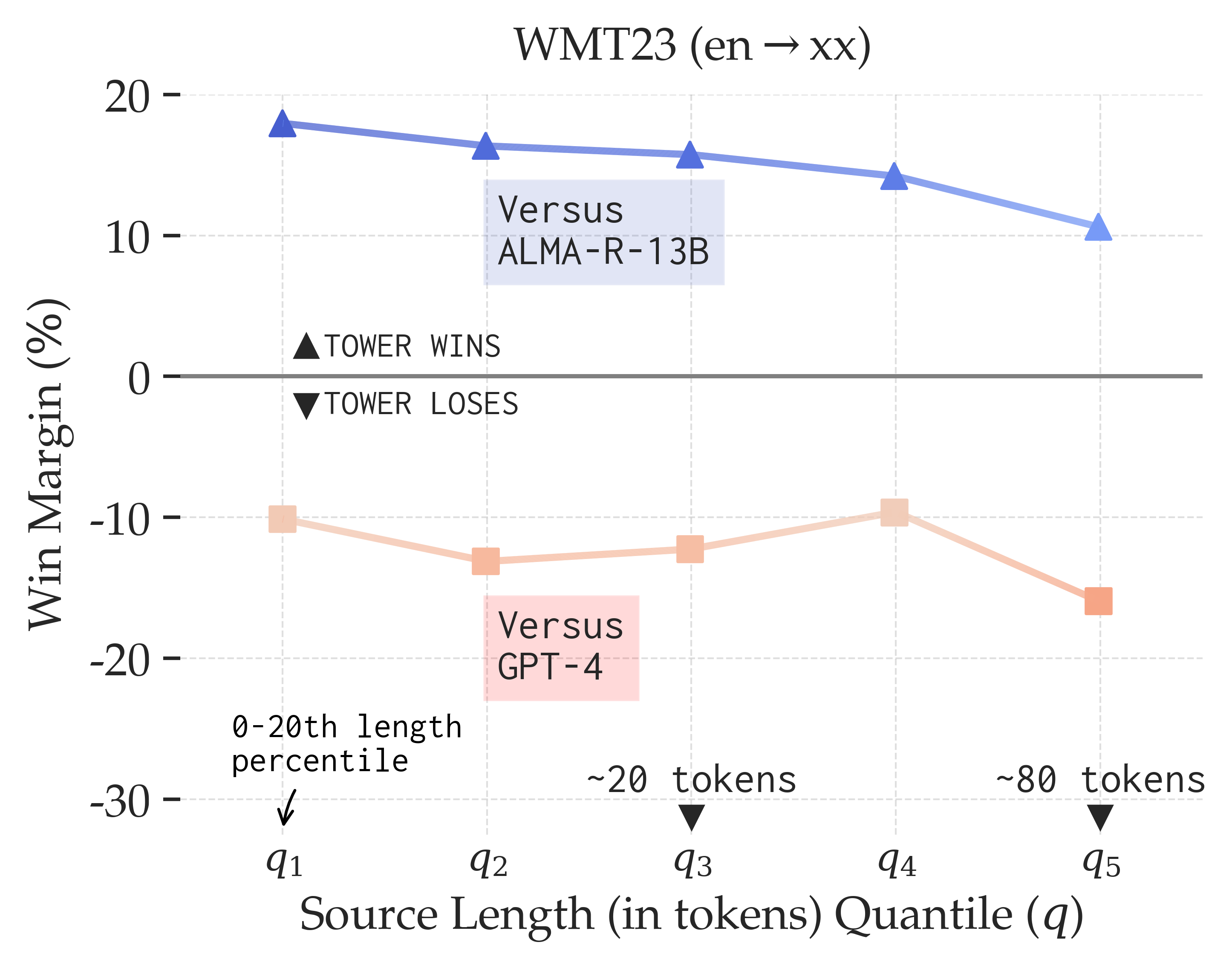}
\caption{en$\rightarrow$xx directions.}
\label{fig:win_margin_en_xx}
\end{subfigure} \quad 
\begin{subfigure}[t]{0.495\textwidth}
\centering
\includegraphics[width=\linewidth]{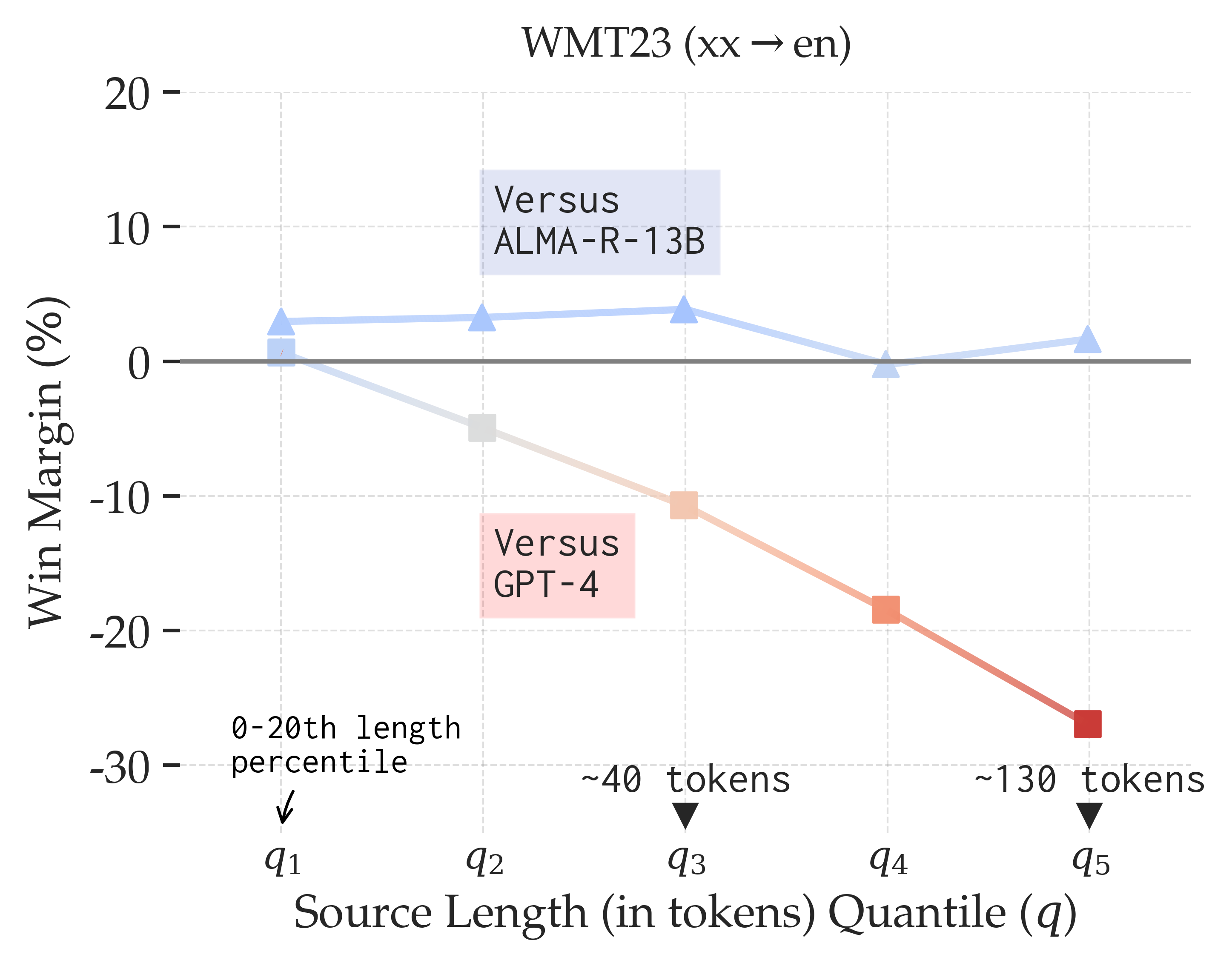}
\caption{xx$\rightarrow$en directions.}
\label{fig:win_margin_xx_en}
\end{subfigure}
\caption{Win rates margin of \TowerInstruct{}-13B by length of the tokenized source for (a) en$\rightarrow$xx and (b) xx$\rightarrow$en language pairs for the \wmt{} test set. We compare against \gptfour{}~($\square$) and \almar{}~($\triangle$). We define a (sentence-level) win if the delta between two systems is superior to 1 \comet{} point.}
\label{fig:win-rate}
\end{figure*}

\begin{table}[t]
\begin{center}
\include{tables/flores_results}
\end{center}
\caption{Translation quality on \flores{} by language pair. Models with statistically significant performance are grouped in quality clusters. Best ranked models are in bold and best ranked open models are underlined.}
\label{tab:flores-results}
\end{table}

We report aggregated results for all models on \flores{}, \wmt{} and \tico{} in Table~\ref{tab:wmt-tico-results}. In Table~\ref{tab:flores-results}, we study the translation quality on all languages in our training set using \flores{}, considering both en$\rightarrow$xx and xx$\rightarrow$en translation directions.

\paragraph{\TowerInstruct{}~13B is the open system with highest translation quality.} \TowerInstruct{}~13B consistently outperforms the larger open models \llama{}~70B and \mixtral{}, as well as the dedicated systems \nllb{}-54B and \almar{} across the board. On \flores{}, \TowerInstruct{}~13B is often ranked first, and is close to \gptfour{} performance on \wmt{} and \tico{}. Upon inspecting both systems' outputs, we verified that the gap between them increases with longer sentences, as is shown in Figure~\ref{fig:win-rate}.\footnote{A similar domain-level analysis did not find any domain dissimilar from the others.} Notably, this trend vanishes when comparing \TowerInstruct{} 13B to \almar{}. We posit this difference stems from a prevalence of shorter sentence-level translations in the training data of both \TowerInstruct{}~13B and \almar{}. In future work, we would like to explore how to better leverage longer contexts, which can benefit instruction-following~\citep{zhao2024long}.

\paragraph{\TowerInstruct{}~13B achieves high translation quality across all language directions.} In Table~\ref{tab:flores-results}, \TowerInstruct{}~13B is ranked first for the majority of en$\rightarrow$xx directions, and is among the top performing models for all but one xx$\rightarrow$en language pair. Notably, \TowerInstruct{} stands out as the best overall model --- outperforming \gptfour{} --- for both pt$\rightarrow$en and ru$\rightarrow$en language pairs. This outcome likely stems from the English-centric pretraining of the \llama{} family. A longer, \textit{more expensive} continued pretraining might improve performance on en$\rightarrow$xx directions further. In fact, we show in Section~\ref{sec:analysis} that the translation quality gains from \llama{} are larger for en$\rightarrow$xx language directions.

\paragraph{\TowerInstruct{} 7B achieves a trade-off between performance and scale.} The smaller \TowerInstruct{} 7B, although behind \TowerInstruct{} 13B, is competitive with other open systems and achieves GPT-3.5-turbo translation quality for some language pairs. Importantly, it outperforms the only system of the same size, \almar{} 7B.

\begin{table}[t]
\begin{center}
\include{tables/final_related_tasks_results}
\end{center}
\caption{Results for translation-related tasks aggregated by language or language pair. Models with statistically significant performance improvements are grouped in quality clusters. We highlight the best ranked models in bold and underline the best ranked \textit{open} models. Since GEC is a held out task, we evaluate all models with 5 in-context examples.}
\label{tab:translation-related-base}
\end{table}

\begin{CJK*}{UTF8}{gbsn}
\begin{figure}[t]
\centering
\includegraphics[width=.9\textwidth]{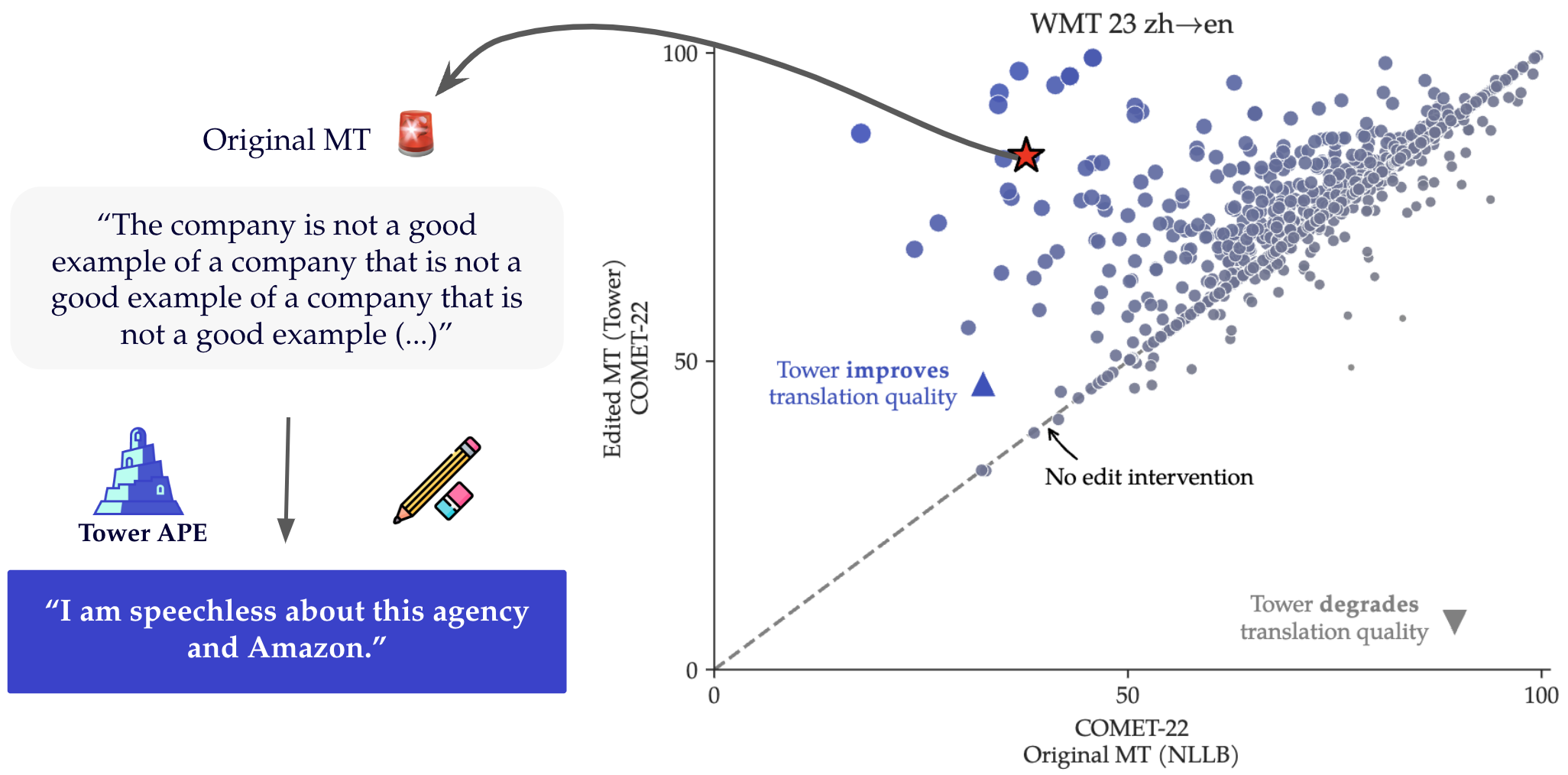}
\caption{Comparison of NLLB 3B original translation quality (x-axis) with \TowerInstruct{} 13B post edition quality (y-axis), and a concrete example (left). Each dot is a WMT 23 zh$\rightarrow$en translation. Marker size and hue represent the difference between post-edition and original translation qualities. The source and reference of the highlighted post edition are ``对这个代理公司和亚马逊实在是很无语。'' and ``As it relates to this agency and Amazon, I am truly stunned.'', respectively. Similar patterns hold on other LPs.}
\label{fig:ape-deep-dive}
\end{figure}
\end{CJK*}

\begin{figure}[t]
\centering
\includegraphics[width=.9\textwidth]{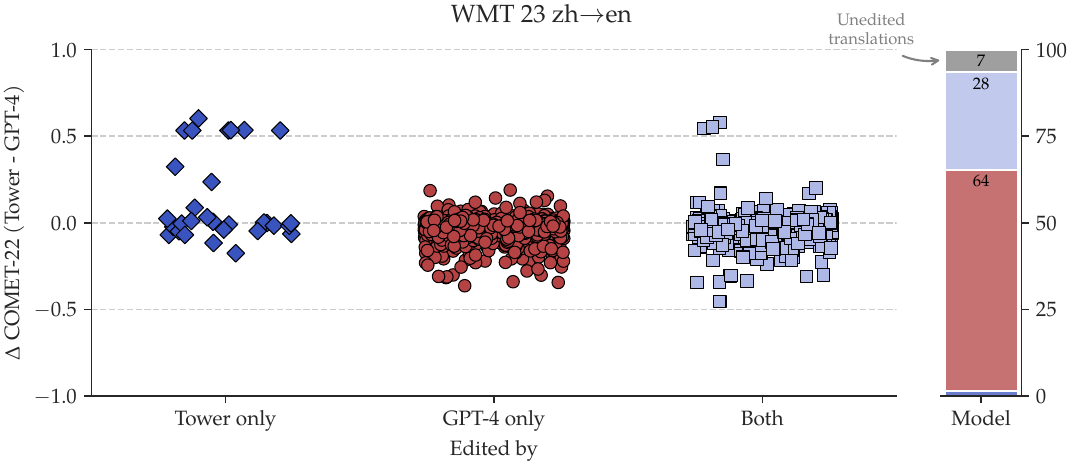}
\caption{Difference in translation quality after post-edition for cases where only \TowerInstruct{} 13B edits ($\diamond$), only \gptfour{} edits ($\circ$), or both models edit (\small{$\square$}). The bar to the right represents the percentage of instances corresponding to each case. Each dot is a \wmt{} zh$\rightarrow$en \nllb{}~3.3B translation, and similar patterns are observed on other LPs.}
\label{fig:tower-gpt-ape}
\end{figure}

\subsection{Translation-Related Tasks}\label{sec:experiments-mt-related}

In Table~\ref{tab:translation-related-base}, we report the results for all translation-related tasks, for both open and closed models, aggregated by language or language pair.\footnote{Appendix~\ref{subsec:appendix-related-tasks-langs} details evaluated languages and provides results for APE and GEC.}

\paragraph{\TowerInstruct{} is an effective translation post editor.} 
\TowerInstruct{} outperforms open models and \gptthreefive{} on APE.
The model's post editions consistently and significantly improve the quality of NLLB 3B translations, going as far as converting oscillatory hallucinations into high-quality translations (Figure~\ref{fig:ape-deep-dive}).
However, \gptfour{} is still the top performer on this task.
One factor that could be behind this gap is that \gptfour{} edits much more often than \TowerInstruct{}, as shown by Figure~\ref{fig:tower-gpt-ape}: almost 90\% of instances are edited by \gptfour{}, compared to the 30\% of \TowerInstruct{}.\footnote{This result suggests that \gptfour{} is over-editing, which we further analyze in Appendix \S\ref{sec:appendix-related-tasks-editing}}
We posit that \TowerInstruct{} learns a tendency for more minimal editing from the relative abundance --- roughly 38\% --- of unedited segments in \TowerBlocks{}.

\paragraph{There is room for improvement on grammatical error correction.} 
On this task, no model significantly outperforms the others on the majority of languages considered.
We hypothesize the relatively average performance of \TowerInstruct{} is caused by the absence of GEC data in  \TowerBlocks{}.

\paragraph{\TowerInstruct{} can identify named entities in multiple languages.}
\TowerInstruct{}~13B shows promising performance on NER, surpassing \gptfour{} by about 15 F1 points.
Similar to APE, most of these improvements are already reflected on \TowerInstruct{} 7B, highlighting its capabilities despite the smaller parameter scale.
Other open models do not perform well on this task, even with 5 in-context examples. 
We hypothesize these results stem from NER being a token-level classification task, as opposed to a generative one. While the models can learn the expected output format from the examples or task description, they struggle to grasp the classification function itself. Conversely, \TowerInstruct{} can learn the task from the records in \TowerBlocks{}.

\input{figs/recipe_ablation}

\section{Dissecting the training recipe}
\label{sec:analysis}

We performed multiple ablations to provide insights on the impact of the several design choices made in the development of the \Tower{} models. 

\paragraph{Continued pretraining and supervised finetuning yield independent performance gains.} The two leftmost plots of Figure~\ref{fig:tower-ablation} illustrate translation quality after continued pretraining and supervised finetuning. Both steps bring performance improvement at both model scales. Remarkably, \TowerBase{} 7B and \TowerInstruct{} 7B outperform \llama{} 13B, and \TowerInstruct{} 7B outperforms \TowerBase{} 13B. In the two rightmost plots, we analyze APE. For this task, while supervised finetuning yields better performance, continued pretraining --- and in particular parallel data --- does not improve performance as observed for translation. In future work, we would like to explore additional training signals during continued pretraining to increase performance for translation-related tasks.

\input{figs/cp_ablation}

\paragraph{Parallel data during continued pretraining improves translation quality.} Figure~\ref{fig:cp-ablation} reports 5-shot translation quality on \flores{} for multiple continued pretraining data recipes. Mixing monolingual and parallel data achieves the highest quality, outperforming both monolingual only and parallel only data. In general, improvements are more noticeable on en$\rightarrow$xx directions, likely due to the English-centric nature of \llama{}'s training. Nevertheless, while monolingual only data improves over the base \llama{} by 0.1 \comet{} points on xx$\rightarrow$en directions, our recipe gains nearly a full point.\footnote{While 0.1 \comet{} points translates to 54.9\% human agreement, one \comet{} point translates to 90.9\%~\citep{kocmi2024navigating}.}

\paragraph{Parallel data during continued pretraining is sample efficient, but quality continues to improve with more tokens.}
At the 2 billion tokens mark, combining parallel sentences with monolingual data (i)~yields more than 50\% of the improvement over the base model, and (ii)~surpasses the recipe leveraging solely monolingual data.
Additionally, while training on more tokens has diminishing returns --- 85\% of the total performance gains appear by the 5 billion tokens mark --- it continues to improve translation quality.

\begin{table}[t]
\begin{center}
\include{tables/sft_ablations}
\end{center}
\caption{Ablation results for the components of \TowerBlocks{}. Results for pretrained models are obtained with 5 in-context examples while results for supervised models are obtained in a 0-shot setup. We consider \flores{} to evaluate translation quality.} 
\label{tab:sft-ablations}
\end{table}

\paragraph{Transfer/interference relations between tasks are complex.} Table~\ref{tab:sft-ablations} ablates the components of \TowerBlocks{}. We finetune on translation data, translation-related tasks including pre- and post-translation, and the full dataset with general-purpose tasks. While adding translation-related tasks improves their performance, it decreases translation quality. We hypothesize that the reduced number of tasks encourages the model to ``split'' its capacity, independently learning each task. Remarkably, introducing general-purpose instructions recovers translation quality, possibly due to the difficulty of ``splitting'' capacity for a large number of tasks.
In future work, we would like to explore transfer/interference between tasks using scaling laws.

\section{Related Work}

Previous work explored various approaches for adapting open models to \textit{single} tasks within the field of machine translation \citep{xu2024alma,xu-etal-2023-instructscore,iyer-etal-2023-towards}, yielding results competitive with closed models or dedicated systems. Notably, \citet{xu2024alma} proposes a two-step approach to adapt \llama{} for translation. Their approach first extends the multilingual capabilities of \llama{} with continued pretraining on \textit{monolingual} data and then specializes for translation by finetuning on high quality parallel data.

Our work also adopts a similar approach, but introduces \textit{parallel} data during continued pretraining and leverages LLMs' instruction-following capabilities to build a system capable of performing \textit{multiple} translation-related tasks.

\paragraph{Multilinguality in LLMs.} While English-centric LLMs can solve tasks in non-English languages, their potential is often limited by the lack of multilingual data in their training corpus.
Works on building more multilingual LLMs bridge this gap in one of two ways: either by training a model ``from scratch'' on more multilingual data~\citep{wei_polylm_2023,Faysse_croissant}, or by continuing the pretraining on data for the language(s) of interest, possibly with vocabulary extension~\citep{chinese-llama-alpaca,xu2024alma,sabia}.

Our multilingual extension approach builds upon insights showcasing the effectiveness of parallel data during pretraining \citep{palm_2,wei_polylm_2023} and includes \textit{parallel} sentences during continued pretraining of \llama{} without vocabulary extension, as preliminary experiments yielded negative results.

\paragraph{Specialization of LLMs.} Recent research also highlights the efficacy of tailoring LLMs for subsets of closely-related tasks.
Again, works are split into training models ``from scratch'' with domain-specific data~\citep{taylor2022galactica,wu2023bloomberggpt}, continued pretraining with data tailored to increase knowledge of the field \citep{lewkowycz2022solving,chen2023meditron70b}, supervised finetuning on domain-specific datasets \citep{yue2024mammoth} or a combination of the last two~\citep{rozière2023codellama,liu2023chipnemo}.

Our specialization approach is broadly inspired by instruction tuning~\citep{wei2022flan,sanh2022t0},\footnote{In this paper, we adopt the nomenclature of supervised finetuning to refer to instruction tuning.} which finetunes language models on a collection of tasks formatted as natural language instructions. Specifically, we curate a dataset for supervised finetuning to specialize LLMs for translation-related tasks. We also leverage the findings from \citet{longpre2022flan,wang-etal-2023-self-instruct,zhou2023lima,xu2024alma}, and prioritize data quality and diversity in our dataset.

\section{Conclusion}

We propose a new recipe for specializing LLMs to \textit{multiple} translation-related tasks. First, we expand the multilingual capabilities of \llama{} with continued pretraining on a highly multilingual corpus. Then, we finetune the model on a dataset of high-quality and diverse instructions for translation-related tasks. Our final model consistently outperforms \textit{open} alternatives on multiple translation-related tasks, and is competitive with \textit{closed-source} models such as GPT-4.

We release the \textsc{Tower} models, as well as \textsc{TowerBlocks}. Moreover, we also make available all the code used for this paper's benchmark, \TowerEval{}, as well as all model generations for the translation benchmark.
The Github repository comes with instructions on how to reproduce the paper's results, and the generations are available on the Zeno platform to allow for interactive exploration.

\subsubsection*{Acknowledgments}
We thank António Farinhas and Manuel Faysse for the fruitful discussion throughout the project. Part of this work was supported by the EU’s Horizon Europe Research and Innovation Actions (UTTER, contract 101070631), by the project DECOLLAGE (ERC-2022-CoG 101088763), by the Portuguese Recovery and Resilience Plan through project C645008882- 00000055 (Center for Responsible AI), and by
Fundação para a Ciência e Tecnologia through contract
UIDB/50008/2020. We also thank GENCI-IDRIS for the technical support and HPC resources --- Jeanzay grants 101838, 103256, 103298 and Adastra grants C1615122, CAD14770, CAD15031 --- used to partially support this work.

\bibliography{tower2024}
\bibliographystyle{tower2024}

\newpage

\input{appendix}
\end{document}

%% file: figs/model_comparison_scale.tex
\definecolor{battleshipgrey}{rgb}{0.3, 0.3, 0.3}
\definecolor{brilliantrose}{rgb}{1.0, 0.33, 0.64}
\definecolor{americanrose}{rgb}{1.0, 0.01, 0.24}
\definecolor{jweigreen}{rgb}{0,0.45,0.24}
\definecolor{bluegray}{rgb}{0.1, 0.1, 0.4}
\definecolor{ao(english)}{rgb}{0.0, 0.5, 0.0}
\definecolor{blanchedalmond}{rgb}{1.0, 0.92, 0.8}
\definecolor{atomictangerine}{rgb}{1.0, 0.6, 0.4}
\definecolor{chocolate(web)}{rgb}{0.82, 0.41, 0.12}
\definecolor{bananayellow}{rgb}{1.0, 0.88, 0.21}
\definecolor{goldenbrown}{rgb}{0.6, 0.4, 0.08}
\definecolor{aliceblue}{rgb}{0.94, 0.97, 1.0}
\definecolor{beige}{rgb}{0.96, 0.96, 0.86}
\definecolor{babyblue}{rgb}{0.54, 0.81, 0.94}
\definecolor{camel}{rgb}{0.76, 0.6, 0.42}
\definecolor{cinnamon}{rgb}{0.82, 0.41, 0.12}
\definecolor{deepskyblue}{rgb}{0.0, 0.75, 1.0}
\definecolor{frenchblue}{rgb}{0.0, 0.45, 0.73}
\definecolor{classicrose}{rgb}{0.98, 0.8, 0.91}
\definecolor{frenchrose}{rgb}{0.96, 0.29, 0.54}
\definecolor{frenchlilac}{rgb}{0.53, 0.38, 0.56}
\definecolor{frenchbeige}{rgb}{0.65, 0.48, 0.36}
\definecolor{metablue}{HTML}{0064E0}
\definecolor{googlegreen}{HTML}{253d7b}
\definecolor{mistralorange}{HTML}{f37004}
\definecolor{openaigreen}{HTML}{10a37f}

\pgfplotsset{
    footnotesize,
    samples=10,
    xmin=1, xmax=82.5,
    ymin=84.9, ymax=89.5,
    xtick={7, 13, 46, 54, 70},
    xticklabels={7, 13, 46, 54, 70},
    ytick={85, 86, 87, 88, 89, 90},
    grid style={dashed, color=gray!30, line width=0.5pt},
    grid=both,
    xlabel={\footnotesize{Model size (\# billion parameters)}},
    x label style={at={(axis description cs:0.5,-0.12)},anchor=north},
}

\begin{figure}[t]
    \begin{centering}
        \begin{tikzpicture}
        \begin{groupplot}[
            group style = {group size = 2 by 1, horizontal sep = 24pt},
            width = 7.5cm, 
            height = 6cm
        ]
            \nextgroupplot[
                align = center,
                title = {\footnotesize{\flores{}}},
                legend style={at={(1.,-0.5)},anchor=south
                ,/tikz/column 2/.style={column sep=8pt,}
                ,/tikz/column 4/.style={column sep=8pt,}
                ,/tikz/column 6/.style={column sep=8pt,}
                ,/tikz/column 8/.style={column sep=8pt,}
                ,/tikz/column 10/.style={column sep=8pt,}
                ,},
                legend cell align={left},
                legend style={draw=none},
                legend columns=3,
                axis x line*=bottom,
                axis y line*=left,
                ylabel={\footnotesize{\comet{}}},
                y label style={at={(axis description cs:-0.07,0.5)},anchor=south},
                xtick pos=bottom,
                ytick pos=left,
            ]

                \addplot[ %
                    color=openaigreen!40, mark size=2pt, dashdotted, line width=1pt,
                ]
                coordinates { (0.91, 88.55) (82.5, 88.55)};

                \addplot[ %
                    color=openaigreen, mark size=2pt, dashdotted, line width=1pt,
                ]
                coordinates {(0.91, 88.78) (82.5, 88.78)};
        
                \addplot[ %
                    color=CustomBlue, fill=CustomBlue!50, mark=triangle*, mark size=4pt, line width=1pt,
                ]
                coordinates {(13, 88.68)};

                \addplot[ %
                    color=CustomBlue, fill=CustomBlue!10, mark=triangle*, mark size=4pt, line width=1pt
                ]
                coordinates {(7, 88.39)};

                \addplot[ %
                    color=metablue, mark=square*, mark size=2.5pt, line width=1pt, fill=metablue!50
                ]
                coordinates { (70, 88.01) };

                \addplot[ %
                    color=metablue, mark=square*, mark size=2.5pt, line width=1pt, fill=metablue!30
                ]
                coordinates { (13, 86.92) };

                \addplot[ %
                    color=metablue, mark=square*, mark size=2.5pt, line width=1pt, fill=metablue!10
                ]
                coordinates { (7, 85.67) };

                \addplot[ %
                    color=mistralorange, mark=pentagon*, mark size=3pt, line width=1pt, fill=mistralorange!50
                ]
                coordinates { (46.7, 87.97) };

                \addplot[ %
                    color=mistralorange, mark=pentagon*, mark size=3pt, line width=1pt, fill=mistralorange!30
                ]
                coordinates { (7, 85.88) };

                \addplot[ %
                    color=metablue, mark=star, mark size=3pt, line width=1pt, fill=metablue!30
                ]
                coordinates { (54, 87.37) };

                \addplot[ %
                    color=googlegreen, mark=*, mark size=3pt, line width=1pt, fill=googlegreen!50
                ]
                coordinates { (7, 87.16) };

                \node at (axis cs:13, 88.68) [anchor=south west, yshift=3pt, xshift=-8pt] {\includegraphics[height=0.3cm]{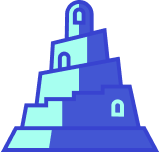} \textcolor{CustomBlue}{\bfseries\scriptsize{\TowerInstruct{}~13B}}};

                \node[anchor=north west, yshift=5pt, xshift=3pt] at (axis cs:7, 88.39) {\textcolor{CustomBlue!70}{\bfseries\scriptsize{\TowerInstruct{}~7B}}};

                \node[anchor=west, xshift=2pt, yshift=2pt] at (axis cs:7, 87.16) {\tiny{\gemma{}~7B}};

                \node[anchor=west, xshift=2.5pt] at (axis cs:13, 86.92) {\tiny{\llama{}~13B}};

                \node[anchor=west, xshift=2.5pt, yshift=-2pt] at (axis cs:7, 85.67) {\tiny{\llama{}~7B}};

                \node[anchor=west, xshift=2.5pt, yshift=2pt] at (axis cs:7, 85.88) {\tiny{\mistral{}}};

                \node[anchor=north, yshift=-1pt] at (axis cs:46.7, 87.97) {\tiny{\mixtral{}}};

                \node[anchor=north, yshift=-1pt] at (axis cs:54, 87.37) {\tiny{\nllb{}~54B}};

                \node[anchor=north, xshift=5pt, yshift=-1pt] at (axis cs:70, 88.01) {\tiny{\llama{}~70B}};

                \node[anchor=north east] at (axis cs:83.75, 88.6) {\bfseries\textcolor{openaigreen!60}{{\tiny{\gptthreefive{}}}}};

                \node[anchor=south east] at (axis cs:83.75, 88.78) {\bfseries\textcolor{openaigreen}{{\tiny{\gptfour{}}}}};

            \nextgroupplot[
                align = center,
                title = {\footnotesize{\wmt{}}},
                axis x line*=bottom,
                axis y line*=left,
                ymin=78, ymax=86,
                ytick={78, 80, 82, 84, 86},
                y label style={at={(axis description cs:-0.05,0.5)},anchor=south},
                xtick pos=bottom,
                ytick pos=left,
            ]

                \addplot[ %
                    color=openaigreen!40, mark size=2pt, dashdotted, line width=1pt,
                ]
                coordinates { (0.91, 84.52) (82.5, 84.52)};
                \node[anchor=north east] at (axis cs:83.75, 84.52) {\bfseries\textcolor{openaigreen!60}{{\tiny{\gptthreefive{}}}}};

                \addplot[ %
                    color=openaigreen, mark size=2pt, dashdotted, line width=1pt,
                ]
                coordinates {(0.91, 84.85) (82.5, 84.85)};

                \node[anchor=south east] at (axis cs:83.75, 84.85) {\bfseries\textcolor{openaigreen}{{\tiny{\gptfour{}}}}};

                \addplot[ %
                    color=CustomBlue, fill=CustomBlue!10, mark=triangle*, mark size=4pt, line width=1pt,
                ]
                coordinates {(7, 83.52)};
                \node[anchor=north west, yshift=2.pt, xshift=1pt] at (axis cs:7, 83.52) {\textcolor{CustomBlue!70}{\bfseries\scriptsize{\TowerInstruct{}-7B}}};

                \addplot[ %
                    color=CustomBlue, fill=CustomBlue!50, mark=triangle*, line width=1pt, mark size = 4pt
                ]
                coordinates {(13, 84.16)};
                \node at (axis cs:13, 84.16) [anchor=west, xshift=2.5pt] {\textcolor{CustomBlue}{\bfseries\scriptsize{\TowerInstruct{}-13B}}};

                \node at (axis cs:13, 84.16) [anchor=south, yshift=3pt] {\includegraphics[height=0.3cm]{figs/Tower_LLM_icon.png}};

                \addplot[ %
                    color=mistralorange, mark=pentagon*, mark size=3pt, line width=1pt, fill=mistralorange!50
                ]
                coordinates { (46.7, 83.22) };
                \node[anchor=north, yshift=-1pt] at (axis cs:46.7, 83.22) {\tiny{\mixtral{}}};

                \addplot[ %
                    color=mistralorange, mark=pentagon*, mark size=3pt, line width=1pt, fill=mistralorange!30
                ]
                coordinates { (7, 80.52) };
                \node[anchor=west, xshift=2.5pt, yshift=2pt] at (axis cs:6, 80.12) {\tiny{\mistral{}}};

                \addplot[ %
                    color=metablue, mark=square*, mark size=2.5pt, line width=1pt, fill=metablue!50
                ]
                coordinates {(70, 82.76)};
                \node[anchor=north, yshift=5pt, xshift=3pt] at (axis cs:70, 82.4) {\textcolor{black}{\tiny{\llama{}~70B}}};

                \addplot[ %
                    color=metablue, mark=square*, mark size=2.5pt, line width=1pt, fill=metablue!30
                ]
                coordinates {(13, 80.70)};
                \node[anchor=west, xshift=2pt] at (axis cs:13, 80.70) {\textcolor{black}{\tiny{\llama{}~13B}}};

                \addplot[ %
                    color=metablue, mark=square*, mark size=2.5pt, line width=1pt, fill=metablue!10
                ]
                coordinates {(7, 78.90)};
                \node[anchor=west, xshift=2pt] at (axis cs:7, 78.90) {\textcolor{black}{\tiny{\llama{}~7B}}};

                \addplot[ %
                    color=metablue, mark=star, mark size=3pt, line width=1pt, fill=metablue!30
                ]
                coordinates {(54, 78.82)};
                \node[anchor=north, yshift=5pt, xshift=3pt] at (axis cs:62, 78.90) {\textcolor{black}{\tiny{\nllb{}~54B}}};

                \addplot[
                    color=darkgray, mark=diamond*, mark size=3pt, line width=1pt, fill=darkgray!30
                ]
                coordinates {(7, 82.89)};
                \node[anchor=west, yshift=-2pt, xshift=1pt] at (axis cs:7, 82.89){\textcolor{black}{\tiny{\almar{}~7B}}};

                \addplot[
                    color=darkgray, mark=diamond*, mark size=3pt, line width=1pt, fill=darkgray!60
                ]
                coordinates {(13, 83.75)};
                \node[anchor=west, xshift=2.5pt, yshift=-0.5pt] at (axis cs:13, 83.75){\textcolor{black}{\tiny{\almar{}~13B}}};

                \addplot[ %
                    color=googlegreen, mark=*, mark size=3pt, line width=1pt, fill=googlegreen!50
                ]
                coordinates { (7, 81.88) };
                \node[anchor=west, xshift=2pt] at (axis cs:7, 81.88) {\tiny{\gemma{}~7B}};
        \end{groupplot}
        \end{tikzpicture}
    \caption{
    Translation quality on \flores{} and \wmt{} for \TowerInstruct{} models and a collection of open and close models across different scales. As the scale of GPT models is not known, we represent them with a horizontal line. \TowerInstruct{} outperforms open alternatives --- even of larger scales --- and is competitive with GPT models.} 
    \label{fig:performance-v-scale}
    \end{centering}
\end{figure}

%% file: tables/translation_results.tex
\renewcommand{\arraystretch}{.9}
\footnotesize
\begin{tabular}{lccccc}
\toprule
 & \multicolumn{2}{c}\flores{} & \multicolumn{2}{c}{WMT 23 } & TICO 19 \\
Models & en$\rightarrow$xx & xx$\rightarrow$en & en$\rightarrow$xx & xx$\rightarrow$en & en$\rightarrow$xx \\
\midrule
\multicolumn{6}{l}{\small \bf Closed} \\
\footnotesize{\gptthreefive{}} & \footnotesize{88.95\secondcluster} & \footnotesize{88.14\thirdcluster} & \footnotesize{85.56\secondcluster} & \footnotesize{83.48\secondcluster} & \footnotesize{87.36\secondcluster} \\
\footnotesize{\gptfour{}} & \footnotesize{\textbf{89.13}\firstcluster} & \footnotesize{\textbf{88.42}\firstcluster} & \footnotesize{\textbf{86.01}\firstcluster} & \footnotesize{\textbf{83.69}\firstcluster} & \footnotesize{\textbf{87.52}\firstcluster} \\
\midrule
\multicolumn{6}{l}{\small \bf Open} \\
\footnotesize{\nllb{} 54B} & \footnotesize{86.79\fourthcluster} & \footnotesize{87.95\thirdcluster} & \footnotesize{78.60\seventhcluster} & \footnotesize{79.06\sixthcluster} & \footnotesize{\underline{87.05}\secondcluster} \\
\footnotesize{\llama{} 70B} & \footnotesize{87.82\fourthcluster} & \footnotesize{88.19\secondcluster} & \footnotesize{82.95\sixthcluster} & \footnotesize{82.56\fourthcluster} & \footnotesize{86.46\fourthcluster} \\
\footnotesize{\mixtral{}} & \footnotesize{87.76\thirdcluster} & \footnotesize{88.17\secondcluster} & \footnotesize{83.60\fifthcluster} & \footnotesize{82.84\thirdcluster} & \footnotesize{86.60\fourthcluster} \\
\footnotesize{\almar{} 7B} & \multicolumn{1}{c}{---} & \multicolumn{1}{c}{---} & \footnotesize{83.40\fifthcluster} & \footnotesize{82.39\fourthcluster} & \multicolumn{1}{c}{---} \\
\footnotesize{\almar{} 13B} & \multicolumn{1}{c}{---} & \multicolumn{1}{c}{---} & \footnotesize{84.46\thirdcluster} & \footnotesize{83.03\thirdcluster} & \multicolumn{1}{c}{---} \\
\cdashlinelr{1-6}
\footnotesize{\TowerInstruct{} 7B} & \footnotesize{88.51\thirdcluster} & \footnotesize{88.27\secondcluster} & \footnotesize{84.28\thirdcluster} & \footnotesize{82.77\fourthcluster} & \footnotesize{87.01\thirdcluster} \\
\footnotesize{\TowerInstruct{} 13B} & \footnotesize{\underline{88.88}\secondcluster} & \footnotesize{\textbf{\underline{88.47}}\firstcluster} & \footnotesize{\underline{85.14}\secondcluster} & \footnotesize{\underline{83.18}\secondcluster} & \footnotesize{\underline{87.32}\secondcluster} \\
\bottomrule
\end{tabular}

%% file: tables/flores_results.tex
\setlength{\tabcolsep}{2.5pt}
\renewcommand{\arraystretch}{.9}
\footnotesize
\begin{tabular}{lccccccccc}
\toprule
 & \multicolumn{9}{c}{\flores{} (en$\rightarrow$xx)} \\
Models & de & es & fr & it & ko & nl & pt & ru & zh \\
\midrule
\multicolumn{10}{l}{\small \bf Closed} \\
\footnotesize{\gptthreefive{}} & \footnotesize{88.78\secondcluster} & \footnotesize{\textbf{87.08}\firstcluster} & \footnotesize{\textbf{89.02}\firstcluster} & \footnotesize{\textbf{89.06}\firstcluster} & \footnotesize{89.36\secondcluster} & \footnotesize{\textbf{88.63}\firstcluster} & \footnotesize{\textbf{90.46}\firstcluster} & \footnotesize{89.56\thirdcluster} & \footnotesize{88.58\secondcluster} \\
\footnotesize{\gptfour{}} & \footnotesize{\textbf{88.98}\firstcluster} & \footnotesize{\textbf{87.10}\firstcluster} & \footnotesize{\textbf{88.93}\firstcluster} & \footnotesize{\textbf{89.05}\firstcluster} & \footnotesize{\textbf{90.06}\firstcluster} & \footnotesize{\textbf{88.56}\firstcluster} & \footnotesize{\textbf{90.43}\firstcluster} & \footnotesize{\textbf{90.19}\firstcluster} & \footnotesize{\textbf{88.87}\firstcluster} \\
\midrule
\multicolumn{10}{l}{\small \bf Open}  \\
\footnotesize{\nllb{} 54B} & \footnotesize{87.18\fifthcluster} & \footnotesize{85.92\fourthcluster} & \footnotesize{87.71\thirdcluster} & \footnotesize{88.10\thirdcluster} & \footnotesize{89.00\thirdcluster} & \footnotesize{87.33\thirdcluster} & \footnotesize{88.72\fifthcluster} & \footnotesize{88.89\fourthcluster} & \footnotesize{78.26\seventhcluster} \\
\footnotesize{\llama{} 70B} & \footnotesize{87.31\fifthcluster} & \footnotesize{86.41\thirdcluster} & \footnotesize{87.82\thirdcluster} & \footnotesize{88.22\thirdcluster} & \footnotesize{88.07\fourthcluster} & \footnotesize{87.47\thirdcluster} & \footnotesize{89.11\fourthcluster} & \footnotesize{88.65\fifthcluster} & \footnotesize{87.32\fifthcluster} \\
\footnotesize{\mixtral{}} & \footnotesize{\underline{87.99}\thirdcluster} & \footnotesize{86.80\secondcluster} & \footnotesize{88.53\secondcluster} & \footnotesize{88.77\secondcluster} & \footnotesize{85.63\fifthcluster} & \footnotesize{87.57\thirdcluster} & \footnotesize{89.45\thirdcluster} & \footnotesize{89.09\fourthcluster} & \footnotesize{85.99\sixthcluster} \\
\cdashlinelr{1-10}
\footnotesize{\TowerInstruct{} 7B} & \footnotesize{87.82\fourthcluster} & \footnotesize{86.76\secondcluster} & \footnotesize{88.44\secondcluster} & \footnotesize{88.73\secondcluster} & \footnotesize{89.41\secondcluster} & \footnotesize{88.38\secondcluster} & \footnotesize{89.60\thirdcluster} & \footnotesize{89.53\thirdcluster} & \footnotesize{87.90\fourthcluster} \\
\footnotesize{\TowerInstruct{} 13B} & \footnotesize{\underline{88.16}\thirdcluster} & \footnotesize{\textbf{\underline{87.06}}\firstcluster} & \footnotesize{\textbf{\underline{88.92}}\firstcluster} & \footnotesize{\textbf{\underline{89.21}}\firstcluster} & \footnotesize{\textbf{\underline{89.92}}\firstcluster} & \footnotesize{\textbf{\underline{88.63}}\firstcluster} & \footnotesize{\underline{89.78}\secondcluster} & \footnotesize{\underline{89.95}\secondcluster} & \footnotesize{\underline{88.29}\thirdcluster} \\
\bottomrule
\\
\toprule
 & \multicolumn{9}{c}{\flores{} (xx$\rightarrow$en)} \\
Models & de & es & fr & it & ko & nl & pt & ru & zh \\
\midrule
\multicolumn{10}{l}{\small \bf Closed}  \\
\footnotesize{\gptthreefive{}} & \footnotesize{89.60\secondcluster} & \footnotesize{87.26\thirdcluster} & \footnotesize{89.46\thirdcluster} & \footnotesize{88.03\thirdcluster} & \footnotesize{87.83\thirdcluster} & \footnotesize{87.71\secondcluster} & \footnotesize{89.78\thirdcluster} & \footnotesize{86.69\fourthcluster} & \footnotesize{86.92\secondcluster} \\
\footnotesize{\gptfour{}} & \footnotesize{\textbf{89.76}\firstcluster} & \footnotesize{\textbf{87.57}\firstcluster} & \footnotesize{\textbf{89.61}\firstcluster} & \footnotesize{88.21\secondcluster} & \footnotesize{\textbf{88.58}\firstcluster} & \footnotesize{\textbf{87.88}\firstcluster} & \footnotesize{89.94\secondcluster} & \footnotesize{86.94\secondcluster} & \footnotesize{\textbf{87.29}\firstcluster} \\
\midrule
\multicolumn{10}{l}{\small \bf Open} \\
\footnotesize{\nllb{} 54B} & \footnotesize{89.17\fourthcluster} & \footnotesize{87.25\thirdcluster} & \footnotesize{89.29\fourthcluster} & \footnotesize{87.91\thirdcluster} & \footnotesize{87.86\thirdcluster} & \footnotesize{87.49\thirdcluster} & \footnotesize{89.38\fourthcluster} & \footnotesize{86.66\fourthcluster} & \footnotesize{86.55\thirdcluster} \\
\footnotesize{\llama{} 70B} & \footnotesize{89.44\thirdcluster} & \footnotesize{87.49\secondcluster} & \footnotesize{89.55\secondcluster} & \footnotesize{88.18\secondcluster} & \footnotesize{87.91\thirdcluster} & \footnotesize{87.52\thirdcluster} & \footnotesize{89.84\secondcluster} & \footnotesize{86.87\secondcluster} & \footnotesize{86.91\secondcluster} \\
\footnotesize{\mixtral{}} & \footnotesize{\underline{89.57}\secondcluster} & \footnotesize{\textbf{\underline{87.65}}\firstcluster} & \footnotesize{89.56\secondcluster} & \footnotesize{\textbf{\underline{88.44}}\firstcluster} & \footnotesize{87.37\fourthcluster} & \footnotesize{87.54\thirdcluster} & \footnotesize{89.73\thirdcluster} & \footnotesize{86.81\thirdcluster} & \footnotesize{86.88\secondcluster} \\
\cdashlinelr{1-10}
\footnotesize{\TowerInstruct{} 7B} & \footnotesize{89.48\thirdcluster} & \footnotesize{87.48\secondcluster} & \footnotesize{89.50\secondcluster} & \footnotesize{\textbf{\underline{88.39}}\firstcluster} & \footnotesize{88.16\secondcluster} & \footnotesize{87.66\secondcluster} & \footnotesize{89.92\secondcluster} & \footnotesize{86.90\secondcluster} & \footnotesize{86.96\secondcluster} \\
\footnotesize{\TowerInstruct{} 13B} & \footnotesize{\underline{89.61}\secondcluster} & \footnotesize{\textbf{\underline{87.62}}\firstcluster} & \footnotesize{\textbf{\underline{89.67}}\firstcluster} & \footnotesize{\textbf{\underline{88.42}}\firstcluster} & \footnotesize{\textbf{\underline{88.48}}\firstcluster} & \footnotesize{\textbf{\underline{87.92}}\firstcluster} & \footnotesize{\textbf{\underline{90.07}}\firstcluster} & \footnotesize{\textbf{\underline{87.20}}\firstcluster} & \footnotesize{\textbf{\underline{87.27}}\firstcluster} \\
\bottomrule
\end{tabular}

%% file: tables/final_related_tasks_results.tex
\renewcommand{\arraystretch}{1}
\footnotesize
\begin{tabular}{lcccc}
\toprule
 & \multicolumn{2}{c}{APE$\uparrow$} & GEC$\downarrow$ & NER$\uparrow$ \\
Models & en$\rightarrow$xx & xx$\rightarrow$en & Multilingual & Multilingual \\
\midrule
\small Baseline (no edits) & \footnotesize{76.80} & \footnotesize{79.99} & \footnotesize{16.66} & \footnotesize{---} \\
\cdashlinelr{1-5}
\multicolumn{5}{l}{\small \bf Closed} \\
\footnotesize{\gptthreefive{}} & \footnotesize{81.47\fourthcluster} & \footnotesize{78.68\fifthcluster} & \footnotesize{15.06\secondcluster} & \footnotesize{50.22\fourthcluster} \\
\footnotesize{\gptfour{}} & \footnotesize{\textbf{85.20}\firstcluster} & \footnotesize{\textbf{84.30}\firstcluster} & \footnotesize{15.08\secondcluster} & \footnotesize{59.88\thirdcluster} \\
\midrule
\multicolumn{5}{l}{\small \bf Open} \\
\footnotesize{\llama{} 70B} & \footnotesize{78.34\fifthcluster} & \footnotesize{81.03\fourthcluster} & \footnotesize{21.74\fifthcluster} & \footnotesize{44.62\fifthcluster} \\
\footnotesize{\mixtral{}} & \footnotesize{82.64\thirdcluster} & \footnotesize{\underline{82.81}\secondcluster} & \footnotesize{17.10\fourthcluster} & \footnotesize{41.77\sixthcluster} \\
\cdashlinelr{1-5}
\footnotesize{\TowerInstruct{} 7B} & \footnotesize{\underline{82.69}\secondcluster} & \footnotesize{81.56\fourthcluster} & \footnotesize{15.13\thirdcluster} & \footnotesize{71.68\secondcluster} \\
\footnotesize{\TowerInstruct{} 13B} & \footnotesize{\underline{83.31}\secondcluster} & \footnotesize{\underline{82.26}\secondcluster} & \footnotesize{\underline{15.68}\secondcluster} & \footnotesize{\textbf{\underline{74.70}}\firstcluster} \\
\bottomrule
\end{tabular}

%% file: figs/recipe_ablation.tex
\definecolor{battleshipgrey}{rgb}{0.3, 0.3, 0.3}
\definecolor{brilliantrose}{rgb}{1.0, 0.33, 0.64}
\definecolor{americanrose}{rgb}{1.0, 0.01, 0.24}
\definecolor{jweigreen}{rgb}{0,0.45,0.24}
\definecolor{bluegray}{rgb}{0.1, 0.1, 0.4}
\definecolor{ao(english)}{rgb}{0.0, 0.5, 0.0}
\definecolor{blanchedalmond}{rgb}{1.0, 0.92, 0.8}
\definecolor{atomictangerine}{rgb}{1.0, 0.6, 0.4}
\definecolor{chocolate(web)}{rgb}{0.82, 0.41, 0.12}
\definecolor{bananayellow}{rgb}{1.0, 0.88, 0.21}
\definecolor{goldenbrown}{rgb}{0.6, 0.4, 0.08}
\definecolor{aliceblue}{rgb}{0.94, 0.97, 1.0}
\definecolor{beige}{rgb}{0.96, 0.96, 0.86}
\definecolor{babyblue}{rgb}{0.54, 0.81, 0.94}
\definecolor{camel}{rgb}{0.76, 0.6, 0.42}
\definecolor{cinnamon}{rgb}{0.82, 0.41, 0.12}
\definecolor{deepskyblue}{rgb}{0.0, 0.75, 1.0}
\definecolor{frenchblue}{rgb}{0.0, 0.45, 0.73}
\definecolor{classicrose}{rgb}{0.98, 0.8, 0.91}
\definecolor{frenchrose}{rgb}{0.96, 0.29, 0.54}
\definecolor{frenchlilac}{rgb}{0.53, 0.38, 0.56}
\definecolor{frenchbeige}{rgb}{0.65, 0.48, 0.36}

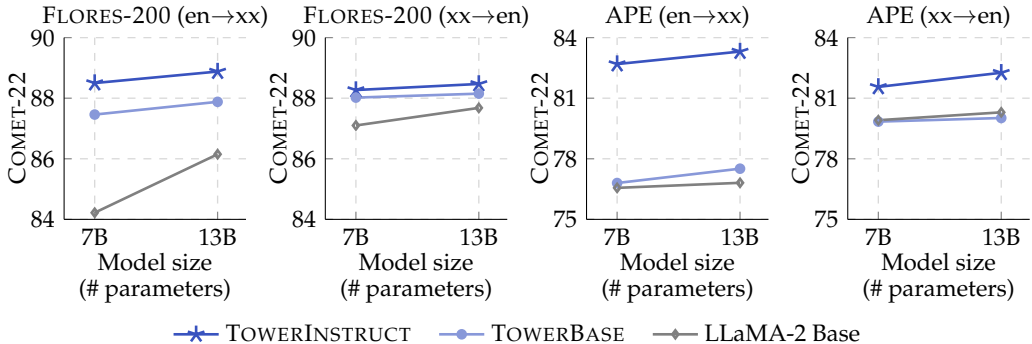
\begin{figure}[t]
    \begin{centering}
    \begin{tikzpicture}
        \pgfplotsset{footnotesize,samples=10}
        \begin{groupplot}[
            group style = {group size = 4 by 1, horizontal sep = 30pt},
            width = 4cm, 
            height = 4cm]
            \nextgroupplot[
                align = center,
                title = {\footnotesize{\flores{} (en$\rightarrow$xx)}},
                legend style={at={(2.5,-0.75)},anchor=south
                ,/tikz/column 2/.style={column sep=8pt,}
                ,/tikz/column 4/.style={column sep=8pt,}
                ,/tikz/column 6/.style={column sep=8pt,}
                ,/tikz/column 8/.style={column sep=8pt,}
                ,/tikz/column 10/.style={column sep=8pt,}
                ,},
                legend style={draw=none},
                legend columns=3,
                xmode=log,
                xmin=6, xmax=15,
                ymin=84, ymax=90,
                xtick={7, 13},
                axis x line*=bottom,
                axis y line*=left,
                xticklabels={7B,13B},
                xlabel={\footnotesize{Model size}\\\footnotesize{(\# parameters)}},
                ylabel={\footnotesize{\comet{}}},
                ytick={80, 82, 84, 86, 88, 90},
                grid style=dashed,
                grid=both,
                x label style={at={(axis description cs:0.5,-0.13)},anchor=north},
                y label style={at={(axis description cs:-0.16,0.5)},anchor=south},
                xtick pos=bottom,
                ytick pos=left,
                ]
                \addplot[ %
                    color=CustomBlue,
                    mark=star,
                    mark size=3.3pt,
                    line width=1pt,
                    ]
                    coordinates {
                    (7, 88.50)
                    (13, 88.88)
                    };
                    \addlegendentry{\textsc{TowerInstruct}}
                \addplot[ %
                    color=CustomBlue!60,
                    mark=*,
                    mark size=1.5pt,
                    line width=1pt,
                    ]
                    coordinates {
                    (7, 87.46)
                    (13, 87.88)
                    };
                    \addlegendentry{\textsc{TowerBase}}
                \addplot[ %
                    color=gray,
                    mark=diamond,
                    mark size=1.5pt,
                    line width=1pt,
                    ]
                    coordinates {
                    (7, 84.22)
                    (13, 86.15)
                    };
                    \addlegendentry{\llama{} Base}
            \nextgroupplot[
                align = center,
                title = {\footnotesize{\flores{} (xx$\rightarrow$en)}},
                legend style={at={(-0.12,1.4)},anchor=south},
                xmode=log,
                xmin=6, xmax=15,
                ymin=84, ymax=90,
                xtick={7, 13},
                axis x line*=bottom,
                axis y line*=left,
                xticklabels={7B,13B},
                xlabel={\footnotesize{Model size}\\\footnotesize{(\# parameters)}},
                ylabel={\footnotesize{\comet{}}},
                ytick={80, 82, 84, 86, 88, 90},
                grid style=dashed,
                grid=both,
                x label style={at={(axis description cs:0.5,-0.13)},anchor=north},
                y label style={at={(axis description cs:-0.16,0.5)},anchor=south},
                xtick pos=bottom,
                ytick pos=left,
                ]
                \addplot[ %
                    color=CustomBlue,
                    mark=star,
                    mark size=3.3pt,
                    line width=1pt,
                    ]
                    coordinates {
                    (7, 88.27)
                    (13, 88.47)
                    };
                \addplot[ %
                    color=CustomBlue!60,
                    mark=*,
                    mark size=1.5pt,
                    line width=1pt,
                    ]
                    coordinates {
                    (7, 88.02)
                    (13, 88.15)
                    };
                \addplot[ %
                    color=gray,
                    mark=diamond,
                    mark size=1.5pt,
                    line width=1pt,
                    ]
                    coordinates {
                    (7, 87.10)
                    (13, 87.68)
                    };
            \nextgroupplot[
                align = center,
                title = {\footnotesize{APE (en$\rightarrow$xx)}},
                legend style={at={(-0.12,1.4)},anchor=south},
                xmode=log,
                xmin=6, xmax=15,
                ymin=75, ymax=84,
                xtick={7, 13},
                axis x line*=bottom,
                axis y line*=left,
                xticklabels={7B,13B},
                xlabel={\footnotesize{Model size}\\\footnotesize{(\# parameters)}},
                ylabel={\footnotesize{\comet{}}},
                ytick={75, 78, 81, 84},
                grid style=dashed,
                grid=both,
                x label style={at={(axis description cs:0.5,-0.13)},anchor=north},
                y label style={at={(axis description cs:-0.16,0.5)},anchor=south},
                xtick pos=bottom,
                ytick pos=left,
                ]
                \addplot[ %
                    color=CustomBlue,
                    mark=star,
                    mark size=3.3pt,
                    line width=1pt,
                    ]
                    coordinates {
                    (7, 82.69)
                    (13, 83.31)
                    };
                \addplot[ %
                    color=CustomBlue!60,
                    mark=*,
                    mark size=1.5pt,
                    line width=1pt,
                    ]
                    coordinates {
                    (7, 76.80)
                    (13, 77.51)
                    };
                \addplot[ %
                    color=gray,
                    mark=diamond,
                    mark size=1.5pt,
                    line width=1pt,
                    ]
                    coordinates {
                    (7, 76.56)
                    (13, 76.81)
                    };
            \nextgroupplot[
                align = center,
                title = {\footnotesize{APE (xx$\rightarrow$en)}},
                legend style={at={(-0.12,1.4)},anchor=south},
                xmode=log,
                xmin=6, xmax=15,
                ymin=75, ymax=84,
                xtick={7, 13},
                axis x line*=bottom,
                axis y line*=left,
                xticklabels={7B,13B},
                xlabel={\footnotesize{Model size}\\\footnotesize{(\# parameters)}},
                ylabel={\footnotesize{\comet{}}},
                ytick={75, 78, 81, 84},
                grid style=dashed,
                grid=both,
                x label style={at={(axis description cs:0.5,-0.13)},anchor=north},
                y label style={at={(axis description cs:-0.16,0.5)},anchor=south},
                xtick pos=bottom,
                ytick pos=left,
                ]
                \addplot[ %
                    color=CustomBlue,
                    mark=star,
                    mark size=3.3pt,
                    line width=1pt,
                    ]
                    coordinates {
                    (7, 81.56)
                    (13, 82.26)
                    };
                \addplot[ %
                    color=CustomBlue!60,
                    mark=*,
                    mark size=1.5pt,
                    line width=1pt,
                    ]
                    coordinates {
                    (7, 79.84)
                    (13, 80.02)
                    };
                \addplot[ %
                    color=gray,
                    mark=diamond,
                    mark size=1.5pt,
                    line width=1pt,
                    ]
                    coordinates {
                    (7, 79.91)
                    (13, 80.30)
                    };
        \end{groupplot}

    \end{tikzpicture}
    \caption{
    Recipe ablation across \textsc{Tower} scales on \flores{} and APE for en$\rightarrow$xx and xx$\rightarrow$en directions. Numbers with pretrained models are obtained in a 5-shot setup; \textsc{TowerInstruct}, on the other hand, is obtained in a 0-shot fashion.
    } 
    \label{fig:tower-ablation}
    \end{centering}
\end{figure}

%% file: figs/cp_ablation.tex
\definecolor{battleshipgrey}{rgb}{0.3, 0.3, 0.3}
\definecolor{brilliantrose}{rgb}{1.0, 0.33, 0.64}
\definecolor{americanrose}{rgb}{1.0, 0.01, 0.24}
\definecolor{jweigreen}{rgb}{0,0.45,0.24}
\definecolor{bluegray}{rgb}{0.1, 0.1, 0.4}
\definecolor{ao(english)}{rgb}{0.0, 0.5, 0.0}
\definecolor{blanchedalmond}{rgb}{1.0, 0.92, 0.8}
\definecolor{atomictangerine}{rgb}{1.0, 0.6, 0.4}
\definecolor{chocolate(web)}{rgb}{0.82, 0.41, 0.12}
\definecolor{bananayellow}{rgb}{1.0, 0.88, 0.21}
\definecolor{goldenbrown}{rgb}{0.6, 0.4, 0.08}
\definecolor{aliceblue}{rgb}{0.94, 0.97, 1.0}
\definecolor{beige}{rgb}{0.96, 0.96, 0.86}
\definecolor{babyblue}{rgb}{0.54, 0.81, 0.94}
\definecolor{camel}{rgb}{0.76, 0.6, 0.42}
\definecolor{cinnamon}{rgb}{0.82, 0.41, 0.12}
\definecolor{deepskyblue}{rgb}{0.0, 0.75, 1.0}
\definecolor{frenchblue}{rgb}{0.0, 0.45, 0.73}
\definecolor{classicrose}{rgb}{0.98, 0.8, 0.91}
\definecolor{frenchrose}{rgb}{0.96, 0.29, 0.54}
\definecolor{frenchlilac}{rgb}{0.53, 0.38, 0.56}
\definecolor{frenchbeige}{rgb}{0.65, 0.48, 0.36}

\pgfplotsset{
    footnotesize,
    samples=10,
    xmin=0.8, xmax=8.1,
    ymin=83.9, ymax=88.5,
    xtick={1,2,3,4,5,6,7,8},
    xticklabels={1,2,3,4,5,6,7,8},
    ytick={84, 85, 86, 87, 88},
    grid style={dashed},
    grid=both,
    xlabel={\footnotesize{Training tokens (billions)}},
    x label style={at={(axis description cs:0.5,-0.12)},anchor=north},
}

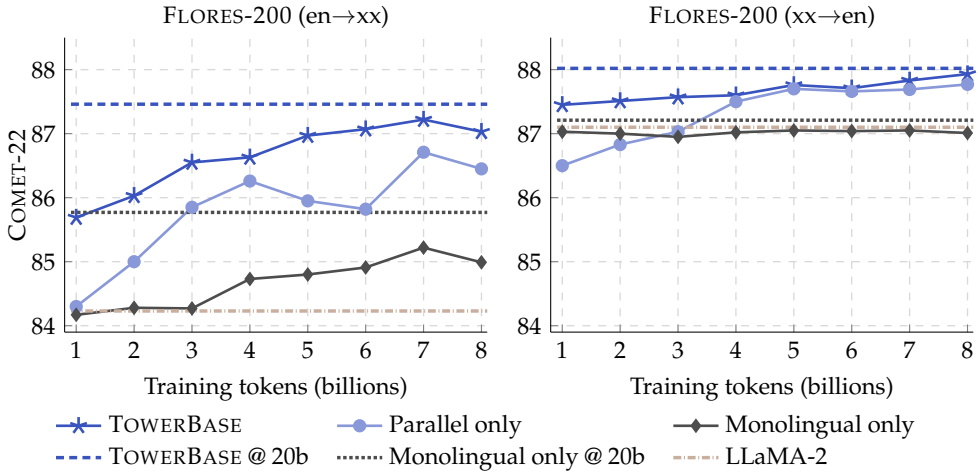
\begin{figure}[t]
    \begin{centering}
        \begin{tikzpicture}
        \begin{groupplot}[
            group style = {group size = 2 by 1, horizontal sep = 24pt},
            width = 7.2cm, 
            height = 5.5cm
        ]
            \nextgroupplot[
                align = center,
                title = {\footnotesize{\flores{} (en$\rightarrow$xx)}},
                legend style={at={(1.,-0.5)},anchor=south
                ,/tikz/column 2/.style={column sep=8pt,}
                ,/tikz/column 4/.style={column sep=8pt,}
                ,/tikz/column 6/.style={column sep=8pt,}
                ,/tikz/column 8/.style={column sep=8pt,}
                ,/tikz/column 10/.style={column sep=8pt,}
                ,},
                legend cell align={left},
                legend style={draw=none},
                legend columns=3,
                axis x line*=bottom,
                axis y line*=left,
                ylabel={\footnotesize{\comet{}}},
                y label style={at={(axis description cs:-0.07,0.5)},anchor=south},
                xtick pos=bottom,
                ytick pos=left,
            ]
                \addplot[ %
                    color=CustomBlue, mark=star, mark size=3.3pt, line width=1pt,
                ]
                coordinates {
    			(1, 85.69) (2, 86.03) (3, 86.55) (4, 86.63)
    			(5, 86.97) (6, 87.07) (7, 87.22) (8, 87.03)
                };
                \addlegendentry{\TowerBase{}}

                \addplot[ %
                    color=CustomBlue!60, mark=*, mark size=2pt, line width=1pt,
                ]
                coordinates {
					(1, 84.3) (2, 85.0) (3, 85.85) (4, 86.26)
                    (5, 85.95) (6, 85.82) (7, 86.71) (8, 86.45)
                };
                \addlegendentry{Parallel only}

                \addplot[ %
                    color=battleshipgrey, mark=diamond*, mark size=2pt, line width=1pt,
                ]
                coordinates {
					(1, 84.17) (2, 84.28) (3, 84.27) (4, 84.73)
					(5, 84.8) (6, 84.91) (7, 85.22) (8, 84.99)
                };
                \addlegendentry{Monolingual only}

                \addplot[ %
                    color=CustomBlue, mark size=3.3pt, densely dashed, line width=1.3pt,
                ]
                coordinates {
                    (0.91, 87.46) (1, 87.46) (2, 87.46) (3, 87.46) (4, 87.46)
                    (5, 87.46) (6, 87.46) (7, 87.46) (8, 87.46) (8.1, 87.46)
                };
                \addlegendentry{\TowerBase{} @ 20b}
                \addplot[ %
                    color=battleshipgrey, mark size=3.3pt, densely dotted, line width=1.3pt,
                ]
                coordinates {
                    (0.91, 85.77) (1, 85.77) (2, 85.77) (3, 85.77) (4, 85.77)
                    (5, 85.77) (6, 85.77) (7, 85.77) (8, 85.77) (8.1, 85.77)
                };
                \addlegendentry{Monolingual only @ 20b}
                \addplot[ %
                    color=frenchbeige!60, mark size=3.3pt, densely dashdotted, line width=1.3pt,
                ]
                coordinates {
                    (0.91, 84.23) (1, 84.23) (2, 84.23) (3, 84.23) (4, 84.23)
					(5, 84.23) (6, 84.23) (7, 84.23) (8, 84.23) (8.1, 84.23)
                };
                \addlegendentry{\llama{}}

            \nextgroupplot[
                align = center,
                title = {\footnotesize{\flores{} (xx$\rightarrow$en)}},
                axis x line*=bottom,
                axis y line*=left,
                y label style={at={(axis description cs:-0.05,0.5)},anchor=south},
                xtick pos=bottom,
                ytick pos=left,
            ]

                \addplot[ %
                    color=CustomBlue, mark=star, mark size=3.3pt, line width=1pt,
                ]
                coordinates {
        		(1, 87.45) (2, 87.51) (3, 87.57) (4, 87.6)
                    (5, 87.76) (6, 87.71) (7, 87.83) (8, 87.93)
                };

                \addplot[ %
                    color=CustomBlue!60, mark=*, mark size=2pt, line width=1pt,
                ]
                coordinates {
					(1, 86.5) (2, 86.83) (3, 87.03) (4, 87.5)
					(5, 87.7) (6, 87.66) (7, 87.69) (8, 87.77)
                };

                \addplot[ %
                    color=battleshipgrey, mark=diamond*, mark size=2pt, line width=1pt,
                ]
                coordinates {
                    (1, 87.03) (2, 87.0) (3, 86.95) (4, 87.02)
                    (5, 87.05) (6, 87.04) (7, 87.05) (8, 87.01)
                };

                \addplot[ %
                    color=CustomBlue, mark size=3.3pt, densely dashed, line width=1.3pt,
                ]
                coordinates {
                    (0.91, 88.02) (1, 88.02) (2, 88.02) (3, 88.02) (4, 88.02)
					(5, 88.02) (6, 88.02) (7, 88.02) (8, 88.02) (8.1, 88.02)
                };
                \addplot[ %
                    color=battleshipgrey, mark size=3.3pt, densely dotted, line width=1.3pt,
                ]
                coordinates {
                    (0.91, 87.21) (1, 87.21) (2, 87.21) (3, 87.21) (4, 87.21)
					(5, 87.21) (6, 87.21) (7, 87.21) (8, 87.21) (8.1, 87.21)
                };
                \addplot[ %
                    color=frenchbeige!60, mark size=3.3pt, densely dashdotted, line width=1.3pt,
                ]
                coordinates {
                    (0.91, 87.10) (1, 87.10) (2, 87.10) (3, 87.10) (4, 87.10)
					(5, 87.10) (6, 87.10) (7, 87.10) (8, 87.10) (8.1, 87.10)
                };
        \end{groupplot}
        \end{tikzpicture}
    \caption{
    Translation quality on \flores{} for continue pretraining data recipes. The \TowerBase{} recipe, outlined in Section~\ref{sec:tower-base}, mixtures monolingual with parallel data. The ``Parallel only'' recipe only processed 8 billion tokens due to compute constraints.} 
    \label{fig:cp-ablation}
    \end{centering}
\end{figure}

%% file: tables/sft_ablations.tex
\renewcommand{\arraystretch}{.9}
\footnotesize
\begin{tabular}{lcccccc}
\toprule
 & \multicolumn{2}{c}{\textbf{MT}} & \multicolumn{2}{c}{\textbf{APE}$\uparrow$} & \textbf{GEC}$\downarrow$ & \textbf{NER}$\uparrow$ \\
Model & en$\rightarrow$xx & xx$\rightarrow$en & en$\rightarrow$xx & xx$\rightarrow$en & Multilingual & Multilingual \\
\midrule
\footnotesize{\llama{} 7B} & 84.23 & 87.10 & 76.56 & 79.91 & 15.95 & 20.09 \\
\footnotesize{\TowerBase{} 7B} & 87.46 & 88.02 & 76.79 & 79.83 & 15.41 & 20.51 \\
\cdashlinelr{1-7}
\multicolumn{7}{l}{\small \bf Supervised Finetuning} \vspace{.5mm}  \\
\footnotesize{+ MT} & 88.45 & \textbf{88.28} & 79.19 & 79.36 & 54.76 & 0.00 \\
\footnotesize{+ Pre-MT + Post-MT} & 87.92 & 87.96 & 81.95 & \textbf{81.73} & 17.44 & \textbf{74.92} \\
\footnotesize{+ General-Purpose} & \textbf{88.51} & 88.27 & \textbf{82.69} & 81.56 & \textbf{15.13} & 71.68 \\
\bottomrule
\end{tabular}

%% file: appendix.tex
\appendix

\section{Analysis of alternative decoding strategies}
\label{sec:appendix-decoding}

\begin{table}[H]
\begin{center}
\input{tables/decoding_strategies}
\end{center}
\caption{Impact of beam search and minimum Bayes risk (MBR) decoding in translation quality for \TowerInstruct{} 13B. In bold, we highlight systems in the first quality cluster. For \tico{} there is no first cluster since no model significantly outperforms the others on a majority of the language pairs.}
\label{tab:decoding-strategies}
\end{table}

In this section, we analyse the performance of \TowerInstruct{}~13B with beam-search~\citep{reddy1977beam} using beam size of 5 and minimum Bayes risk (MBR) decoding~\citep{eikema-aziz-2020-map,fernandes-etal-2022-quality,freitag-etal-2022-high} with 20 hypotheses and \comet{} as an utility function. We generate hypotheses using temperature and nucleus sampling~\citep{Holtzman2020The}, with $t = 0.9$ and $p = 0.6$. We avoid ``optimizing'' the evaluation metric \citep{fernandes-etal-2022-quality} by measuring translation quality with \bleurt{}.

Table~\ref{tab:decoding-strategies} reports translation quality across all test sets. Both decoding strategies consistently improve translation quality over greedy decoding, with MBR decoding achieving higher quality. Additionally, for both \wmt{} and \tico{}, decoding strategies close the gap to \gptfour{}. Notably, on \flores{}, \TowerInstruct{} 13B appears isolated in the first cluster.

\section{Further analysis on \TowerInstruct{} and \gptfour{} editing tendencies}\label{sec:appendix-related-tasks-editing}

\begin{figure}[h]
    \centering
    \includegraphics[width=0.5\textwidth]{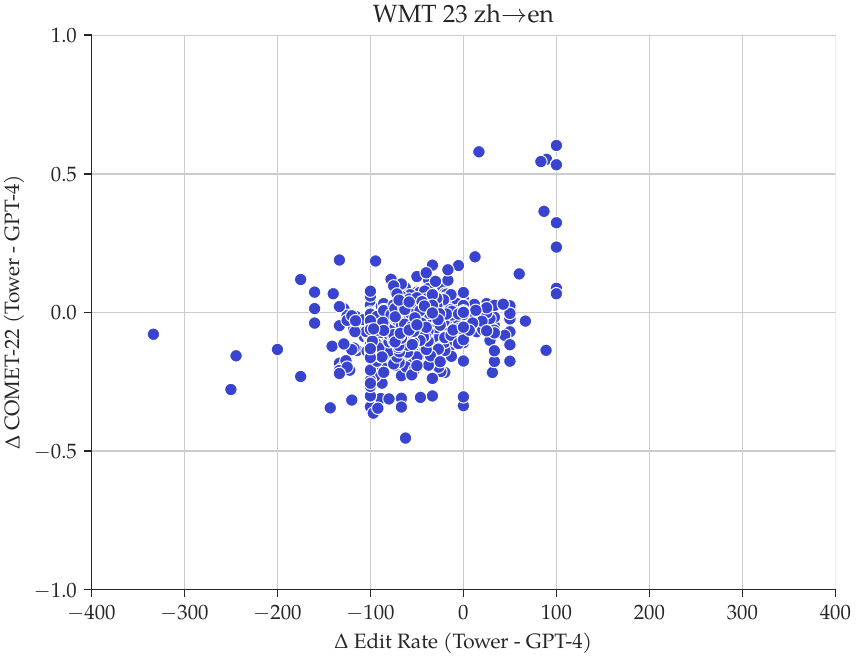}
    \caption{Difference between \TowerInstruct{} 13B and \gptfour{} edit rate (compared to the original NLLB translation) (x-axis), and difference between \TowerInstruct{} 13B and \gptfour{} post-edition \comet{} (y-axis). The correlation between the two variables is 0.34 Spearman $\rho$. Similar patterns are observed for other language pairs.}
    \label{fig:appendix-ter-comparison}
\end{figure}

Figure~\ref{fig:appendix-ter-comparison} shows that differences between \gptfour{} and \TowerInstruct{} edit rates are not strongly correlated to differences in \comet{} (0.34 Spearman $\rho$). 
This means that \gptfour{} edits often do not correspond to gains in performance.
This finding, allied with the discussion in Section~\ref{sec:experiments-mt-related} about \gptfour{} editing considerably more than \TowerInstruct{}, suggests that \gptfour{} may be editing too much.

\section{Details of the continued pretraining dataset}
\label{sec:appendix-cp-data}
In Table \ref{tab:cp-quality-thresholds}, we report the perplexity floors and ceilings used to filter the monolingual data in the continued pretraining corpus, as well as the Bicleaner and CometKiwi-22 thresholds used to filter the parallel data. In Table \ref{tab:parallel_data}, we also detail all sources of the parallel sentences used in the continued pretraining dataset.

\begin{table}[h]
\centering \small
\setlength{\tabcolsep}{2.35ex}
\begin{tabular}{@{}lccccccccc@{}}
\toprule
                    & en  & de   & fr   & nl   & es   & pt   & ru   & zh   & ko   \\ \midrule
Min. perplexity $\ast$    & 50  & 50   & 50   & 50   & 50   & 50   & 50   & 50   & 50   \\
Max. perplexity $\ast$  & 516 & 611  & 322  & 649  & 275  & 257  & 334  & 2041 & 198  \\
Bicleaner $\dagger$ & -   & 0.5  & 0.5  & 0.5  & 0.5  & 0.5  & 0.5  & 0.0  & 0.5  \\
\cometkiwi{} $\dagger$  & -   & 0.75 & 0.75 & 0.75 & 0.75 & 0.75 & 0.75 & 0.75 & 0.75 \\ \bottomrule
\end{tabular}
\caption{Quality filtering thresholds applied on monolingual data ($\ast$) and parallel data ($\dagger$) by language. On the latter, the to-English language pair's threshold is the same as the corresponding from-English one.}
\label{tab:cp-quality-thresholds}
\end{table}

\begin{table}[h]
\centering \small
\setlength{\tabcolsep}{2.35ex}
\begin{tabular}{lccc}
\toprule
Dataset         & Version   &  \\ \midrule
Europarl \citep{koehn-2005-europarl}      & v8                 &  \\
ParaCrawl \citep{espla-etal-2019-paracrawl}       & v9                  &  \\
MultiParaCrawl \citep{espla-etal-2019-paracrawl}  & v7.1               &  \\
CCMatrix \citep{schwenk2020ccmatrix}       & v1                &  \\
CCAligned \citep{el-kishky-etal-2020-ccaligned}      & v1               &  \\
MultiCCAligned \citep{el-kishky-etal-2020-ccaligned} & v1              &  \\
WikiTitles \citep{tiedemann2012opus}     & v2014           &  \\
WikiMatrix \citep{schwenk2019wikimatrix}     & v1              &  \\
News-Commentary \citep{tiedemann2012opus} & v16                 &  \\
OPUS100 \citep{zhang2020improving} & v1                 &  \\
TildeModel \citep{rozis-skadins-2017-tilde}     & v2018          &  \\
Bible \citep{mayer-cysouw-2014-creating}          & v1               &  \\
Ubuntu \citep{tiedemann2012opus}         & v14.10           &  \\
Tatoeba \citep{tiedemann2012opus}        & v2            &  \\
GNOME \citep{tiedemann2012opus}          & v1          &  \\
GlobalVoices  \citep{tiedemann2012opus}  & v2018q4         &  \\
KDE4  \citep{tiedemann2012opus}          & v2               &  \\
KDE-Doc  \citep{tiedemann2012opus}       & v1             &  \\
PHP \citep{tiedemann2012opus}            & v1           &  \\
Wikipedia \citep{Wo_k_2014}      & v1.0             &  \\
Wikimedia \citep{tiedemann2012opus}      & v20210402       &  \\
JRC \citep{tiedemann2012opus}            & v3.0           &  \\
DGT \citep{tiedemann2012opus}            & v2019              &  \\
EuroPat \citep{europat}        & v3              &  \\
EUbookshop \citep{tiedemann2012opus}     & v2                &  \\
EMEA \citep{tiedemann2012opus}           & v3                &  \\
EUConst \citep{tiedemann2012opus}        & v1               &  \\
tico-19 \citep{anastasopoulos-etal-2020-tico}        & v20201028        &  \\
ECB \citep{tiedemann2012opus}            & v1                 &  \\
Elitr-ECA \citep{williams2021elitr}      & v1            &  \\
MultiUN \citep{eisele-chen-2010-multiun}        & v1               &  \\
OpenOffice \citep{tiedemann2012opus}     & v3                 &  \\
Ada83 \citep{tiedemann2012opus}          & v1                &  \\
infopankki \citep{tiedemann2012opus}     & v1          &  \\
Scielo \citep{soares2018large}         & v1               &  \\
giga-fren \citep{tiedemann2012opus}      & v2               &  \\
UNPC \citep{ziemski-etal-2016-united}           & v1.0           & \\ \bottomrule
\end{tabular} 
\caption{The various data sources used to create the parallel data with the number of available language pairs.}
\label{tab:parallel_data}
\end{table}

\section{Details of \TowerBlocks{}}
\label{sec:tower-blocks-details}

This appendix details all datasets utilized in \TowerBlocks{}:
\begin{itemize}
    \item \textbf{WMT14 to WMT21}\footnote{\url{https://www2.statmt.org/wmt23/translation-task.html}} --- Evaluation sets for the general machine translation shared task;
    \item \textbf{WMT22 with quality-shots}~\citep{hendy2023gptmt} --- Evaluation set from WMT23 with high quality in-context examples;
    \item \textbf{NTREX}~\citep{federmann-etal-2022-ntrex} --- Professional translations of the WMT19 test set;
    \item \textbf{\flores{}}~\citep{nllbteam2022} --- Development set of the \flores{} dataset for all languages included in training;
    \item \textbf{FRMT}~\citep{riley2022frmt} --- Human translations of English Wikipedia sentences into regional variants;
    \item \textbf{OPUS}~\citep{tiedemann2012opus} --- Parallel corpora from which we sampled very high-quality samples for all language pairs;
    \item \textbf{QT21}~\citep{specia2017qt21} and \textbf{ApeQuest}\footnote{\url{https://apequest.wordpress.com/}} --- Translation data with post-edits utilized for general translation and automatic post-editing;
    \item \textbf{MT-GenEval}~\citep{currey-etal-2022-mtgeneval} --- Gender translation benchmark which we leveraged for general translation and context-aware translation;
    \item \textbf{WMT20 to WMT22 Metrics MQM}\footnote{\url{https://www.statmt.org/wmt22/results.html}} --- MT evaluation data annotated with multidimensional quality metrics~\citep{Lommel_Burchardt_Uszkoreit_2014mqm} that we used to perform error span detection;
    \item \textbf{WMT17 to WMT22 Metrics DAs}\footnote{\url{https://www.statmt.org/wmt22/results.html}} --- MT evaluation data annotated with direct assessements (DAs)~\citep{graham2013da} which we utilized for translation ranking.
    \item \textbf{WMT21 Terminology}\footnote{\url{https://www.statmt.org/wmt21/terminology-task.html}} --- Development set for the WMT21 terminology task;
    \item \textbf{Tatoeba}~\citep{tiedemann-2020-tatoeba} --- Development set of the Tatoeba dataset which we used to generate translations in different languages for the same source --- we identified this task as multi-reference translation;
    \item \textbf{MultiCoNER 2022 and 2023}~\citep{malmasi-etal-2022-multiconer,fetahu-etal-2023-multiconer} --- Development sets of the named entity recognition MultiCoNER datasets. For MultiCoNER 2023, we adopted the coarse-grained entity categorization;
    \item \textbf{PAWS-X}~\citep{pawsx2019emnlp} --- Development set of the PAWS-X dataset which we used as paraphrase generation;
    \item \textbf{UltraChat}~\citep{ding2023ultrachat} --- Filtered version of the UltraChat dataset used in \citet{tunstall2023zephyr};
    \item \textbf{Glaive Code Assistant}\footnote{\url{https://huggingface.co/datasets/glaiveai/glaive-code-assistant}} --- Coding questions and answers across a wide range of programming languages.
\end{itemize}

\clearpage

\section{Details of \TowerInstruct{}}

\subsection{Hyperparameters}
\label{sec:appendix-tower-instruct-hyperparams}
Table~\ref{tab:tower-instruct-hyperparams} details the full hyperparameters configuration for the training of \TowerInstruct{}. We also utilized bfloat16 mixed precision and packing.

\begin{table}[ht]
\renewcommand{\arraystretch}{1.1}
\begin{center}
\begin{tabular}{ll}
\toprule
Global train batch size & 256 \\
Number of Epochs & 4 \\
Learning rate & 7e-6 \\
LR Scheduler & cosine \\
Warmup Steps & 500 \\
Weight Decay & 0.01 \\
Optimizer & Adam \citep{Kingma2014AdamAM} \\
Adam $\beta_1$ & 0.9 \\
Adam $\beta_2$ & 0.999 \\
Adam $\epsilon$ & 1e-8 \\
Maximum Sequence Length & 2048 \\
\bottomrule
\end{tabular}    
\end{center}
\caption{Hyperparameter configuration to finetune \TowerInstruct{} on \TowerBlocks{}.}
\label{tab:tower-instruct-hyperparams}
\end{table}

\subsection{Chat Template}
\label{sec:appendix-chat-template}

We finetuned \TowerInstruct{} with the \texttt{chatml} template~\citep{chatml}. Table~\ref{tab:dialogue_example} provides an example of an interaction using the aforementioned template.

\begin{table}[ht]
\centering
\footnotesize
\begin{tabular}{p{1cm}p{12cm}}
\toprule
\textbf{User}   & {\textcolor{CustomBlue}{\texttt{<|im\_start|>user}}} \\
                 & Translate the following text from Portuguese into English. \\
                 & Portuguese: Ontem, a minha amiga foi ao supermercado mas estava fechado. Queria comprar legumes e fruta. \\
                 & English: {\textcolor{CustomBlue}{\texttt{<|im\_end|>}}} \\
                 & {\textcolor{CustomBlue}{\texttt{<|im\_start|>assistant}}} \\
\textbf{Model}   & Yesteday, my friend went to the supermarket but it was closed. She wanted to buy vegetables and fruit.{\textcolor{CustomBlue}{\texttt{<|im\_end|>}}}\\
\textbf{User}   & {\textcolor{CustomBlue}{\texttt{<|im\_start|>user}}} \\
                 & Can you now translate it into Spanish? {\textcolor{CustomBlue}{\texttt{<|im\_end|>}}} \\
                 & {\textcolor{CustomBlue}{\texttt{<|im\_start|>assistant}}} \\
\textbf{Model}   & Ayer mi amiga fue al supermercado, pero estaba cerrado. Quería comprar verduras y fruta.{\textcolor{CustomBlue}{\texttt{<|im\_end|>}}}\\
\bottomrule
\end{tabular}
\caption{Example of a dialogue with \TowerInstruct{}'s user and model control tokens.}
\label{tab:dialogue_example}
\end{table}

\clearpage

\section{Translation full results}\label{sec:appendix-translation}

On Tables~\ref{tab:wmt-tico-results-xcomet} to Tables~\ref{tab:wmt-tico-results-chrf}, we tables equivalent to Table~\ref{tab:wmt-tico-results}, but with different metrics (one per table): \xcomet{}, \cometkiwi{}, \bleurt{}, and \chrf{}.
The equivalent for Table~\ref{tab:flores-results} is done in Tables~\ref{tab:flores-results-xcomet} to~\ref{tab:flores-results-chrf}.
On Tables~\ref{tab:flores-everything},~\ref{tab:wmt-everything}, and~\ref{tab:tico-everything}, we present translation results for a wider variety of models, broken down by language pair.

\begin{table}[h]
\begin{center}
\include{tables/appendix_wmt_tico_xcomet}
\end{center}
\caption{Translation quality on \wmt{} and \tico{} by language pair measured by \xcomet{}. Models with statistically significant performance are grouped in quality clusters. Best performing models are in bold and best performing open models are underlined.}
\label{tab:wmt-tico-results-xcomet}
\end{table}

\begin{table}[h]
\begin{center}
\include{tables/appendix_wmt_tico_kiwi}
\end{center}
\caption{Translation quality on \wmt{} and \tico{} by language pair measured by \cometkiwi{}. Models with statistically significant performance are grouped in quality clusters. Best performing models are in bold and best performing open models are underlined.}
\label{tab:wmt-tico-results-kiwi}
\end{table}

\begin{table}[h]
\begin{center}
\include{tables/appendix_wmt_tico_bleurt}
\end{center}
\caption{Translation quality on \wmt{} and \tico{} by language pair measured by \bleurt{}. Models with statistically significant performance are grouped in quality clusters. Best performing models are in bold and best performing open models are underlined.}
\label{tab:wmt-tico-results-bleurt}
\end{table}

\begin{table}[h]
\begin{center}
\include{tables/appendix_wmt_tico_chrf}
\end{center}
\caption{Translation quality on \wmt{} and \tico{} by language pair measured by \chrf{}. Models with statistically significant performance are grouped in quality clusters. Best performing models are in bold and best performing open models are underlined.}
\label{tab:wmt-tico-results-chrf}
\end{table}

\begin{table}[h]
\begin{center}
\include{tables/appendix_flores_xcomet}
\end{center}
\caption{Translation quality on \flores{} by language pair measured by \xcomet{}. Models with statistically significant performance are grouped in quality clusters. Best performing models are in bold and best performing open models are underlined.}
\label{tab:flores-results-xcomet}
\end{table}

\begin{table}[h]
\begin{center}
\include{tables/appendix_flores_kiwi}
\end{center}
\caption{Translation quality on \flores{} by language pair measured by \cometkiwi{}. Models with statistically significant performance are grouped in quality clusters. Best performing models are in bold and best performing open models are underlined.}
\label{tab:flores-results-kiwi}
\end{table}

\begin{table}[h]
\begin{center}
\include{tables/appendix_flores_bleurt}
\end{center}
\caption{Translation quality on \flores{} by language pair measured by \bleurt{}. Models with statistically significant performance are grouped in quality clusters. Best performing models are in bold and best performing open models are underlined.}
\label{tab:flores-results-bleurt}
\end{table}

\begin{table}[h]
\begin{center}
\include{tables/appendix_flores_chrf}
\end{center}
\caption{Translation quality on \flores{} by language pair measured by \chrf{}. Models with statistically significant performance are grouped in quality clusters. Best performing models are in bold and best performing open models are underlined.}
\label{tab:flores-results-chrf}
\end{table}

\begin{table}[h]
\begin{center}
\input{tables/flores_everything}
\end{center}
\caption{\comet{} on \flores{} for a wide variety of models.}
\label{tab:flores-everything}
\end{table}

\begin{table}[h]
\begin{center}
\input{tables/wmt_everything}
\end{center}
\caption{\comet{} on \wmt{} for a wide variety of models.}
\label{tab:wmt-everything}
\end{table}

\begin{table}[h]
\begin{center}
\input{tables/tico_everything}
\end{center}
\caption{\comet{} on \tico{} for a wide variety of models.}
\label{tab:tico-everything}
\end{table}

\clearpage

\section{Translation-related tasks full results}\label{sec:appendix-related-tasks}

\subsection{Languages considered}\label{subsec:appendix-related-tasks-langs}

For APE, on Table~\ref{tab:translation-related-base}, we consider 4 language pairs: en$\rightarrow$de, en$\rightarrow$zh, de$\rightarrow$en, and ru$\rightarrow$en. 
We leave out en$\rightarrow$ru and zh$\rightarrow$en, because we had no post editions to serve as fewshot examples for \llama{} and \mixtral{}.
In any case, we provide results for \TowerInstruct{}, \gptthreefive{}, and \gptfour{} on the 6 language pairs in Table~\ref{tab:appendix-ape-full}.

For NER, we consider English, German, French, Spanish, Italian, Portuguese, Russian, and Chinese.
Finally, we evaluate GEC on English, German, and Spanish. For this task, besides the numbers shown in Table~\ref{tab:translation-related-base}, we also measure \errant{} in Table~\ref{tab:appendix-gec}.

Results broken down by language may be found in Tables~\ref{tab:appendix-ape-everything},~\ref{tab:appendix-gec-everything}, and~\ref{tab:appendix-ner-everything}.

\begin{table}[h]
\begin{center}
\input{tables/appendix_ape}
\end{center}
\caption{APE results for the 6 WMT23 LPs considered. \nllb{} corresponds to the translations that were subject to editing, so their quality serves as the baseline for the task. Table~\ref{tab:translation-related-base} did not include zh-en and en-ru to guarantee a fair comparison with open models --- there were no fewshot examples available for these LPs.}
\label{tab:appendix-ape-full}
\end{table}

\begin{table}[h]
\begin{center}
\input{tables/appendix_gec}
\end{center}
\caption{GEC \errant{} results.}
\label{tab:appendix-gec}
\end{table}

\begin{table}[h]
\begin{center}
\input{tables/appendix_ape_everything}
\end{center}
\caption{APE \comet{} results by language pair.}
\label{tab:appendix-ape-everything}
\end{table}

\begin{table}[h]
\begin{center}
\input{tables/appendix_gec_everything}
\end{center}
\caption{GEC edit rate results by language.}
\label{tab:appendix-gec-everything}
\end{table}

\begin{table}[h]
\begin{center}
\input{tables/appendix_ner_everything}
\end{center}
\caption{NER F1 results by language.}
\label{tab:appendix-ner-everything}
\end{table}

%% file: tables/decoding_strategies.tex
\setlength{\tabcolsep}{2.5pt}
\renewcommand{\arraystretch}{1}
\footnotesize
\begin{tabular}{lccccc}
\toprule
 & \multicolumn{2}{c}\flores{} & \multicolumn{2}{c}{WMT 23 } & TICO 19 \\
Models & en$\rightarrow$xx & xx$\rightarrow$en & en$\rightarrow$xx & xx$\rightarrow$en & en$\rightarrow$xx \\
\midrule
\footnotesize{\gptthreefive{}} & \footnotesize{77.08} & \footnotesize{78.12} & \footnotesize{72.06} & \footnotesize{\textbf{72.50}} & \footnotesize{75.91} \\
\footnotesize{\gptfour{}} & \footnotesize{77.26} & \footnotesize{78.51} & \footnotesize{\textbf{72.54}} & \footnotesize{\textbf{72.91}} & \footnotesize{\textbf{76.16}} \\
\midrule
\multicolumn{6}{l}{\small \TowerInstruct{} 13B} \vspace{.5mm}  \\
\footnotesize{Greedy} & \footnotesize{76.89} & \footnotesize{78.67} & \footnotesize{70.87} & \footnotesize{71.75} & \footnotesize{75.40} \\
\footnotesize{Beam} & \footnotesize{77.40} & \footnotesize{\textbf{78.87}} & \footnotesize{71.31} & \footnotesize{71.88} & \footnotesize{75.66} \\
\footnotesize{MBR} & \footnotesize{\textbf{77.79}} & \footnotesize{\textbf{78.96}} & \footnotesize{72.29} & \footnotesize{72.36} & \footnotesize{76.13} \\
\bottomrule
\end{tabular}

%% file: tables/appendix_wmt_tico_xcomet.tex
\renewcommand{\arraystretch}{1}
\footnotesize
\begin{tabular}{lccccc}
\toprule
 & \multicolumn{2}{c}\flores{} & \multicolumn{2}{c}{WMT 23 } & TICO 19 \\
Models & en$\rightarrow$xx & xx$\rightarrow$en & en$\rightarrow$xx & xx$\rightarrow$en & en$\rightarrow$xx \\
\midrule
\multicolumn{6}{l}{\small \bf Closed} \vspace{.5mm}  \\
\footnotesize{\gptthreefive{}} & \footnotesize{94.41\secondcluster} & \footnotesize{\textbf{95.54}\firstcluster} & \footnotesize{88.99\secondcluster} & \footnotesize{89.75\secondcluster} & \footnotesize{91.19\secondcluster} \\
\footnotesize{\gptfour{}} & \footnotesize{\textbf{94.75}\firstcluster} & \footnotesize{\textbf{96.01}\firstcluster} & \footnotesize{\textbf{89.46}\firstcluster} & \footnotesize{\textbf{90.28}\firstcluster} & \footnotesize{91.38\secondcluster} \\
\midrule
\multicolumn{6}{l}{\small \bf Open} \vspace{.5mm}  \\
\footnotesize{\nllb{} 54B} & \footnotesize{90.04\fourthcluster} & \footnotesize{93.78\fourthcluster} & \footnotesize{78.99\sixthcluster} & \footnotesize{81.38\sixthcluster} & \footnotesize{90.11\thirdcluster} \\
\footnotesize{\llama{} 70B} & \footnotesize{92.80\fourthcluster} & \footnotesize{94.15\fourthcluster} & \footnotesize{84.85\sixthcluster} & \footnotesize{87.21\fifthcluster} & \footnotesize{89.02\fifthcluster} \\
\footnotesize{\mixtral{}} & \footnotesize{91.90\thirdcluster} & \footnotesize{94.40\thirdcluster} & \footnotesize{85.67\sixthcluster} & \footnotesize{87.81\fourthcluster} & \footnotesize{89.30\fourthcluster} \\
\footnotesize{\almar{} 7B} & \multicolumn{1}{c}{---} & \multicolumn{1}{c}{---} & \footnotesize{86.50\fourthcluster} & \footnotesize{87.67\fourthcluster} & \multicolumn{1}{c}{---} \\
\footnotesize{\almar{} 13B} & \multicolumn{1}{c}{---} & \multicolumn{1}{c}{---} & \footnotesize{\underline{88.88}\secondcluster} & \footnotesize{\underline{88.97}\thirdcluster} & \multicolumn{1}{c}{---} \\
\cdashlinelr{1-6}
\footnotesize{\TowerInstruct{} 7B} & \footnotesize{93.85\secondcluster} & \footnotesize{94.67\thirdcluster} & \footnotesize{87.20\fourthcluster} & \footnotesize{87.88\fourthcluster} & \footnotesize{90.56\thirdcluster} \\
\footnotesize{\TowerInstruct{} 13B} & \footnotesize{\textbf{\underline{94.80}}\firstcluster} & \footnotesize{\underline{95.22}\secondcluster} & \footnotesize{\underline{88.71}\secondcluster} & \footnotesize{\underline{88.65}\thirdcluster} & \footnotesize{\underline{91.30}\secondcluster} \\
\bottomrule
\end{tabular}

%% file: tables/appendix_wmt_tico_kiwi.tex
\renewcommand{\arraystretch}{1}
\footnotesize
\begin{tabular}{lccccc}
\toprule
 & \multicolumn{2}{c}\flores{} & \multicolumn{2}{c}{WMT 23 } & TICO 19 \\
Models & en$\rightarrow$xx & xx$\rightarrow$en & en$\rightarrow$xx & xx$\rightarrow$en & en$\rightarrow$xx \\
\midrule
\multicolumn{6}{l}{\small \bf Closed} \vspace{.5mm}  \\
\footnotesize{\gptthreefive{}} & \footnotesize{86.25\secondcluster} & \footnotesize{85.64\secondcluster} & \footnotesize{80.82\secondcluster} & \footnotesize{80.35\secondcluster} & \footnotesize{85.65\secondcluster} \\
\footnotesize{\gptfour{}} & \footnotesize{\textbf{86.42}\firstcluster} & \footnotesize{\textbf{85.77}\firstcluster} & \footnotesize{\textbf{81.20}\firstcluster} & \footnotesize{\textbf{80.54}\firstcluster} & \footnotesize{85.79\secondcluster} \\
\midrule
\multicolumn{6}{l}{\small \bf Open} \vspace{.5mm}  \\
\footnotesize{\nllb{} 54B} & \footnotesize{82.93\fifthcluster} & \footnotesize{84.89\fourthcluster} & \footnotesize{70.96\sixthcluster} & \footnotesize{76.69\fifthcluster} & \footnotesize{85.16\thirdcluster} \\
\footnotesize{\llama{} 70B} & \footnotesize{85.30\fourthcluster} & \footnotesize{84.97\fourthcluster} & \footnotesize{78.43\fifthcluster} & \footnotesize{79.36\fourthcluster} & \footnotesize{84.66\fifthcluster} \\
\footnotesize{\mixtral{}} & \footnotesize{85.24\thirdcluster} & \footnotesize{85.32\thirdcluster} & \footnotesize{79.01\fifthcluster} & \footnotesize{79.82\thirdcluster} & \footnotesize{84.81\fourthcluster} \\
\footnotesize{\almar{} 7B} & \multicolumn{1}{c}{---} & \multicolumn{1}{c}{---} & \footnotesize{79.25\fourthcluster} & \footnotesize{79.79\fourthcluster} & \multicolumn{1}{c}{---} \\
\footnotesize{\almar{} 13B} & \multicolumn{1}{c}{---} & \multicolumn{1}{c}{---} & \footnotesize{80.12\thirdcluster} & \footnotesize{\underline{80.21}\secondcluster} & \multicolumn{1}{c}{---} \\
\cdashlinelr{1-6}
\footnotesize{\TowerInstruct{} 7B} & \footnotesize{85.96\thirdcluster} & \footnotesize{85.41\thirdcluster} & \footnotesize{79.80\fourthcluster} & \footnotesize{79.95\thirdcluster} & \footnotesize{85.32\thirdcluster} \\
\footnotesize{\TowerInstruct{} 13B} & \footnotesize{\underline{86.19}\secondcluster} & \footnotesize{\underline{85.51}\secondcluster} & \footnotesize{\underline{80.57}\secondcluster} & \footnotesize{\underline{80.25}\secondcluster} & \footnotesize{\underline{85.59}\secondcluster} \\
\bottomrule
\end{tabular}

%% file: tables/appendix_wmt_tico_bleurt.tex
\renewcommand{\arraystretch}{1}
\footnotesize
\begin{tabular}{lccccc}
\toprule
 & \multicolumn{2}{c}\flores{} & \multicolumn{2}{c}{WMT 23 } & TICO 19 \\
Models & en$\rightarrow$xx & xx$\rightarrow$en & en$\rightarrow$xx & xx$\rightarrow$en & en$\rightarrow$xx \\
\midrule
\multicolumn{6}{l}{\small \bf Closed} \vspace{.5mm}  \\
\footnotesize{\gptthreefive{}} & \footnotesize{\textbf{77.08}\firstcluster} & \footnotesize{78.12\thirdcluster} & \footnotesize{72.06\secondcluster} & \footnotesize{\textbf{72.50}\firstcluster} & \footnotesize{75.91\secondcluster} \\
\footnotesize{\gptfour{}} & \footnotesize{\textbf{77.26}\firstcluster} & \footnotesize{78.51\secondcluster} & \footnotesize{\textbf{72.54}\firstcluster} & \footnotesize{\textbf{72.91}\firstcluster} & \footnotesize{76.16\secondcluster} \\
\midrule
\multicolumn{6}{l}{\small \bf Open} \vspace{.5mm}  \\
\footnotesize{\nllb{} 54B} & \footnotesize{74.29\thirdcluster} & \footnotesize{77.99\thirdcluster} & \footnotesize{62.73\sixthcluster} & \footnotesize{66.46\fifthcluster} & \footnotesize{\underline{75.49}\secondcluster} \\
\footnotesize{\llama{} 70B} & \footnotesize{75.04\fourthcluster} & \footnotesize{78.28\secondcluster} & \footnotesize{68.03\fifthcluster} & \footnotesize{71.01\thirdcluster} & \footnotesize{74.00\fourthcluster} \\
\footnotesize{\mixtral{}} & \footnotesize{74.78\thirdcluster} & \footnotesize{78.10\secondcluster} & \footnotesize{68.81\fifthcluster} & \footnotesize{71.32\thirdcluster} & \footnotesize{74.22\fourthcluster} \\
\footnotesize{\almar{} 7B} & \multicolumn{1}{c}{---} & \multicolumn{1}{c}{---} & \footnotesize{68.64\fifthcluster} & \footnotesize{70.66\fourthcluster} & \multicolumn{1}{c}{---} \\
\footnotesize{\almar{} 13B} & \multicolumn{1}{c}{---} & \multicolumn{1}{c}{---} & \footnotesize{70.09\fourthcluster} & \footnotesize{71.47\thirdcluster} & \multicolumn{1}{c}{---} \\
\cdashlinelr{1-6}
\footnotesize{\TowerInstruct{} 7B} & \footnotesize{76.10\thirdcluster} & \footnotesize{78.26\secondcluster} & \footnotesize{69.77\fourthcluster} & \footnotesize{71.11\thirdcluster} & \footnotesize{74.83\fourthcluster} \\
\footnotesize{\TowerInstruct{} 13B} & \footnotesize{\underline{76.89}\secondcluster} & \footnotesize{\textbf{\underline{78.67}}\firstcluster} & \footnotesize{\underline{70.87}\secondcluster} & \footnotesize{\underline{71.75}\secondcluster} & \footnotesize{75.40\thirdcluster} \\
\bottomrule
\end{tabular}

%% file: tables/appendix_wmt_tico_chrf.tex
\renewcommand{\arraystretch}{1}
\footnotesize
\begin{tabular}{lccccc}
\toprule
 & \multicolumn{2}{c}\flores{} & \multicolumn{2}{c}{WMT 23 } & TICO 19 \\
Models & en$\rightarrow$xx & xx$\rightarrow$en & en$\rightarrow$xx & xx$\rightarrow$en & en$\rightarrow$xx \\
\midrule
\multicolumn{6}{l}{\small \bf Closed} \vspace{.5mm}  \\
\footnotesize{\gptthreefive{}} & \footnotesize{\textbf{58.20}\firstcluster} & \footnotesize{63.75\thirdcluster} & \footnotesize{\textbf{56.38}\firstcluster} & \footnotesize{60.92\secondcluster} & \footnotesize{64.18\secondcluster} \\
\footnotesize{\gptfour{}} & \footnotesize{\textbf{58.61}\firstcluster} & \footnotesize{64.35\secondcluster} & \footnotesize{\textbf{56.94}\firstcluster} & \footnotesize{\textbf{61.33}\firstcluster} & \footnotesize{64.34\secondcluster} \\
\midrule
\multicolumn{6}{l}{\small \bf Open} \vspace{.5mm}  \\
\footnotesize{\nllb{} 54B} & \footnotesize{54.70\fourthcluster} & \footnotesize{63.87\secondcluster} & \footnotesize{42.98\sixthcluster} & \footnotesize{52.08\sixthcluster} & \footnotesize{\underline{63.84}\secondcluster} \\
\footnotesize{\llama{} 70B} & \footnotesize{55.19\fourthcluster} & \footnotesize{64.15\secondcluster} & \footnotesize{52.31\fourthcluster} & \footnotesize{\underline{59.66}\secondcluster} & \footnotesize{61.65\fourthcluster} \\
\footnotesize{\mixtral{}} & \footnotesize{54.50\fourthcluster} & \footnotesize{63.38\thirdcluster} & \footnotesize{51.22\fourthcluster} & \footnotesize{58.63\fourthcluster} & \footnotesize{61.34\fourthcluster} \\
\footnotesize{\almar{} 7B} & \multicolumn{1}{c}{---} & \multicolumn{1}{c}{---} & \footnotesize{45.20\seventhcluster} & \footnotesize{57.33\fourthcluster} & \multicolumn{1}{c}{---} \\
\footnotesize{\almar{} 13B} & \multicolumn{1}{c}{---} & \multicolumn{1}{c}{---} & \footnotesize{46.52\sixthcluster} & \footnotesize{58.37\thirdcluster} & \multicolumn{1}{c}{---} \\
\cdashlinelr{1-6}
\footnotesize{\TowerInstruct{} 7B} & \footnotesize{56.16\thirdcluster} & \footnotesize{64.08\secondcluster} & \footnotesize{52.25\fourthcluster} & \footnotesize{58.88\fourthcluster} & \footnotesize{62.07\fourthcluster} \\
\footnotesize{\TowerInstruct{} 13B} & \footnotesize{\underline{57.19}\secondcluster} & \footnotesize{\textbf{\underline{64.79}}\firstcluster} & \footnotesize{\underline{54.10}\thirdcluster} & \footnotesize{\underline{59.78}\secondcluster} & \footnotesize{62.81\thirdcluster} \\
\bottomrule
\end{tabular}

%% file: tables/appendix_flores_xcomet.tex
\renewcommand{\arraystretch}{1}
\footnotesize
\begin{tabular}{lccccc}
\toprule
 & \multicolumn{2}{c}\flores{} & \multicolumn{2}{c}{WMT 23 } & TICO 19 \\
Models & en$\rightarrow$xx & xx$\rightarrow$en & en$\rightarrow$xx & xx$\rightarrow$en & en$\rightarrow$xx \\
\midrule
\multicolumn{6}{l}{\small \bf Closed} \vspace{.5mm}  \\
\footnotesize{\gptthreefive{}} & \footnotesize{94.41\secondcluster} & \footnotesize{\textbf{95.54}\firstcluster} & \footnotesize{88.99\secondcluster} & \footnotesize{89.75\secondcluster} & \footnotesize{91.19\secondcluster} \\
\footnotesize{\gptfour{}} & \footnotesize{\textbf{94.75}\firstcluster} & \footnotesize{\textbf{96.01}\firstcluster} & \footnotesize{\textbf{89.46}\firstcluster} & \footnotesize{\textbf{90.28}\firstcluster} & \footnotesize{91.38\secondcluster} \\
\midrule
\multicolumn{6}{l}{\small \bf Open} \vspace{.5mm}  \\
\footnotesize{\nllb{} 54B} & \footnotesize{90.04\fourthcluster} & \footnotesize{93.78\fourthcluster} & \footnotesize{78.99\sixthcluster} & \footnotesize{81.38\sixthcluster} & \footnotesize{90.11\thirdcluster} \\
\footnotesize{\llama{} 70B} & \footnotesize{92.80\fourthcluster} & \footnotesize{94.15\fourthcluster} & \footnotesize{84.85\sixthcluster} & \footnotesize{87.21\fifthcluster} & \footnotesize{89.02\fifthcluster} \\
\footnotesize{\mixtral{}} & \footnotesize{91.90\thirdcluster} & \footnotesize{94.40\thirdcluster} & \footnotesize{85.67\sixthcluster} & \footnotesize{87.81\fourthcluster} & \footnotesize{89.30\fourthcluster} \\
\footnotesize{\almar{} 7B} & \multicolumn{1}{c}{---} & \multicolumn{1}{c}{---} & \footnotesize{86.50\fourthcluster} & \footnotesize{87.67\fourthcluster} & \multicolumn{1}{c}{---} \\
\footnotesize{\almar{} 13B} & \multicolumn{1}{c}{---} & \multicolumn{1}{c}{---} & \footnotesize{\underline{88.88}\secondcluster} & \footnotesize{\underline{88.97}\thirdcluster} & \multicolumn{1}{c}{---} \\
\cdashlinelr{1-6}
\footnotesize{\TowerInstruct{} 7B} & \footnotesize{93.85\secondcluster} & \footnotesize{94.67\thirdcluster} & \footnotesize{87.20\fourthcluster} & \footnotesize{87.88\fourthcluster} & \footnotesize{90.56\thirdcluster} \\
\footnotesize{\TowerInstruct{} 13B} & \footnotesize{\textbf{\underline{94.80}}\firstcluster} & \footnotesize{\underline{95.22}\secondcluster} & \footnotesize{\underline{88.71}\secondcluster} & \footnotesize{\underline{88.65}\thirdcluster} & \footnotesize{\underline{91.30}\secondcluster} \\
\bottomrule
\end{tabular}

%% file: tables/appendix_flores_kiwi.tex
\setlength{\tabcolsep}{2.5pt}
\renewcommand{\arraystretch}{1}
\footnotesize
\begin{tabular}{lccccccccc}
\toprule
 & \multicolumn{9}{c}{\flores{} (en$\rightarrow$xx)} \\
Models & de & es & fr & it & ko & nl & pt & ru & zh \\
\midrule
\multicolumn{10}{l}{\small \bf Closed} \vspace{.5mm}  \\
\footnotesize{\gptthreefive{}} & \footnotesize{85.15\secondcluster} & \footnotesize{\textbf{87.04}\firstcluster} & \footnotesize{\textbf{87.18}\firstcluster} & \footnotesize{\textbf{87.47}\firstcluster} & \footnotesize{86.92\thirdcluster} & \footnotesize{\textbf{86.88}\firstcluster} & \footnotesize{85.69\secondcluster} & \footnotesize{85.58\secondcluster} & \footnotesize{84.37\secondcluster} \\
\footnotesize{\gptfour{}} & \footnotesize{\textbf{85.27}\firstcluster} & \footnotesize{\textbf{87.07}\firstcluster} & \footnotesize{\textbf{87.25}\firstcluster} & \footnotesize{\textbf{87.51}\firstcluster} & \footnotesize{\textbf{87.47}\firstcluster} & \footnotesize{\textbf{86.90}\firstcluster} & \footnotesize{85.68\secondcluster} & \footnotesize{\textbf{85.99}\firstcluster} & \footnotesize{\textbf{84.68}\firstcluster} \\
\midrule
\multicolumn{10}{l}{\small \bf Open} \vspace{.5mm}  \\
\footnotesize{\nllb{} 54B} & \footnotesize{82.59\sixthcluster} & \footnotesize{85.18\fourthcluster} & \footnotesize{85.23\fourthcluster} & \footnotesize{85.66\fourthcluster} & \footnotesize{86.11\fourthcluster} & \footnotesize{84.71\fourthcluster} & \footnotesize{83.45\fifthcluster} & \footnotesize{83.56\fourthcluster} & \footnotesize{69.88\seventhcluster} \\
\footnotesize{\llama{} 70B} & \footnotesize{84.19\fifthcluster} & \footnotesize{86.40\thirdcluster} & \footnotesize{86.68\thirdcluster} & \footnotesize{86.77\thirdcluster} & \footnotesize{85.46\fifthcluster} & \footnotesize{85.87\thirdcluster} & \footnotesize{84.57\fourthcluster} & \footnotesize{84.59\thirdcluster} & \footnotesize{83.13\fifthcluster} \\
\footnotesize{\mixtral{}} & \footnotesize{\underline{84.72}\thirdcluster} & \footnotesize{86.74\secondcluster} & \footnotesize{87.04\secondcluster} & \footnotesize{87.18\secondcluster} & \footnotesize{83.49\sixthcluster} & \footnotesize{85.95\thirdcluster} & \footnotesize{84.99\thirdcluster} & \footnotesize{84.78\thirdcluster} & \footnotesize{82.30\sixthcluster} \\
\cdashlinelr{1-10}
\footnotesize{\TowerInstruct{} 7B} & \footnotesize{84.41\fourthcluster} & \footnotesize{86.77\secondcluster} & \footnotesize{87.08\secondcluster} & \footnotesize{87.31\secondcluster} & \footnotesize{86.70\thirdcluster} & \footnotesize{\underline{86.48}\secondcluster} & \footnotesize{85.57\secondcluster} & \footnotesize{\underline{85.50}\secondcluster} & \footnotesize{83.78\fourthcluster} \\
\footnotesize{\TowerInstruct{} 13B} & \footnotesize{\underline{84.73}\thirdcluster} & \footnotesize{\textbf{\underline{86.94}}\firstcluster} & \footnotesize{\textbf{\underline{87.18}}\firstcluster} & \footnotesize{\textbf{\underline{87.45}}\firstcluster} & \footnotesize{\underline{87.22}\secondcluster} & \footnotesize{\underline{86.60}\secondcluster} & \footnotesize{\textbf{\underline{85.85}}\firstcluster} & \footnotesize{\underline{85.68}\secondcluster} & \footnotesize{\underline{84.09}\thirdcluster} \\
\bottomrule
\\
\toprule
 & \multicolumn{9}{c}{\flores{} (xx$\rightarrow$en)} \\
Models & de & es & fr & it & ko & nl & pt & ru & zh \\
\midrule
\multicolumn{10}{l}{\small \bf Closed} \vspace{.5mm}  \\
\footnotesize{\gptthreefive{}} & \footnotesize{84.64\secondcluster} & \footnotesize{86.27\secondcluster} & \footnotesize{\textbf{86.48}\firstcluster} & \footnotesize{86.84\secondcluster} & \footnotesize{85.69\secondcluster} & \footnotesize{86.18\secondcluster} & \footnotesize{\textbf{85.31}\firstcluster} & \footnotesize{84.59\secondcluster} & \footnotesize{84.76\secondcluster} \\
\footnotesize{\gptfour{}} & \footnotesize{\textbf{84.71}\firstcluster} & \footnotesize{\textbf{86.39}\firstcluster} & \footnotesize{\textbf{86.50}\firstcluster} & \footnotesize{\textbf{86.95}\firstcluster} & \footnotesize{\textbf{86.15}\firstcluster} & \footnotesize{\textbf{86.25}\firstcluster} & \footnotesize{\textbf{85.31}\firstcluster} & \footnotesize{\textbf{84.75}\firstcluster} & \footnotesize{\textbf{84.92}\firstcluster} \\
\midrule
\multicolumn{10}{l}{\small \bf Open} \vspace{.5mm}  \\
\footnotesize{\nllb{} 54B} & \footnotesize{84.09\fifthcluster} & \footnotesize{85.51\fifthcluster} & \footnotesize{86.04\thirdcluster} & \footnotesize{86.06\fourthcluster} & \footnotesize{85.13\fourthcluster} & \footnotesize{85.59\fifthcluster} & \footnotesize{84.45\fourthcluster} & \footnotesize{83.95\fourthcluster} & \footnotesize{83.18\sixthcluster} \\
\footnotesize{\llama{} 70B} & \footnotesize{84.29\fourthcluster} & \footnotesize{85.78\fourthcluster} & \footnotesize{86.05\thirdcluster} & \footnotesize{86.38\thirdcluster} & \footnotesize{84.45\sixthcluster} & \footnotesize{85.56\fifthcluster} & \footnotesize{84.87\thirdcluster} & \footnotesize{83.77\fourthcluster} & \footnotesize{83.57\fifthcluster} \\
\footnotesize{\mixtral{}} & \footnotesize{\underline{84.45}\thirdcluster} & \footnotesize{\underline{86.07}\thirdcluster} & \footnotesize{\underline{86.34}\secondcluster} & \footnotesize{\underline{86.78}\secondcluster} & \footnotesize{84.74\fifthcluster} & \footnotesize{85.78\fourthcluster} & \footnotesize{\underline{85.13}\secondcluster} & \footnotesize{84.45\thirdcluster} & \footnotesize{84.14\fourthcluster} \\
\cdashlinelr{1-10}
\footnotesize{\TowerInstruct{} 7B} & \footnotesize{\underline{84.41}\thirdcluster} & \footnotesize{\underline{86.12}\thirdcluster} & \footnotesize{\underline{86.35}\secondcluster} & \footnotesize{\underline{86.79}\secondcluster} & \footnotesize{85.21\fourthcluster} & \footnotesize{\underline{85.98}\thirdcluster} & \footnotesize{\underline{85.17}\secondcluster} & \footnotesize{84.47\secondcluster} & \footnotesize{84.16\fourthcluster} \\
\footnotesize{\TowerInstruct{} 13B} & \footnotesize{\underline{84.44}\thirdcluster} & \footnotesize{\underline{86.09}\thirdcluster} & \footnotesize{\underline{86.39}\secondcluster} & \footnotesize{\underline{86.83}\secondcluster} & \footnotesize{\underline{85.47}\thirdcluster} & \footnotesize{\underline{86.04}\thirdcluster} & \footnotesize{\underline{85.17}\secondcluster} & \footnotesize{\textbf{\underline{84.69}}\firstcluster} & \footnotesize{\underline{84.47}\thirdcluster} \\
\bottomrule
\end{tabular}

%% file: tables/appendix_flores_bleurt.tex
\setlength{\tabcolsep}{2.5pt}
\renewcommand{\arraystretch}{1}
\footnotesize
\begin{tabular}{lccccccccc}
\toprule
 & \multicolumn{9}{c}{\flores{} (en$\rightarrow$xx)} \\
Models & de & es & fr & it & ko & nl & pt & ru & zh \\
\midrule
\multicolumn{10}{l}{\small \bf Closed} \vspace{.5mm}  \\
\footnotesize{\gptthreefive{}} & \footnotesize{\textbf{79.09}\firstcluster} & \footnotesize{\textbf{76.75}\firstcluster} & \footnotesize{\textbf{79.54}\firstcluster} & \footnotesize{79.83\secondcluster} & \footnotesize{69.39\secondcluster} & \footnotesize{\textbf{77.79}\firstcluster} & \footnotesize{\textbf{80.31}\firstcluster} & \footnotesize{77.31\secondcluster} & \footnotesize{73.69\secondcluster} \\
\footnotesize{\gptfour{}} & \footnotesize{\textbf{79.13}\firstcluster} & \footnotesize{\textbf{76.64}\firstcluster} & \footnotesize{\textbf{79.29}\firstcluster} & \footnotesize{80.00\secondcluster} & \footnotesize{\textbf{70.31}\firstcluster} & \footnotesize{77.58\secondcluster} & \footnotesize{\textbf{80.22}\firstcluster} & \footnotesize{\textbf{78.16}\firstcluster} & \footnotesize{\textbf{73.98}\firstcluster} \\
\midrule
\multicolumn{10}{l}{\small \bf Open} \vspace{.5mm}  \\
\footnotesize{\nllb{} 54B} & \footnotesize{77.71\thirdcluster} & \footnotesize{75.37\fourthcluster} & \footnotesize{77.96\thirdcluster} & \footnotesize{79.26\thirdcluster} & \footnotesize{68.95\secondcluster} & \footnotesize{76.47\thirdcluster} & \footnotesize{77.80\fourthcluster} & \footnotesize{76.81\thirdcluster} & \footnotesize{58.32\sixthcluster} \\
\footnotesize{\llama{} 70B} & \footnotesize{76.75\fourthcluster} & \footnotesize{75.28\fifthcluster} & \footnotesize{76.96\fourthcluster} & \footnotesize{78.70\fourthcluster} & \footnotesize{67.01\thirdcluster} & \footnotesize{75.98\fourthcluster} & \footnotesize{77.50\fourthcluster} & \footnotesize{75.79\fourthcluster} & \footnotesize{71.41\fourthcluster} \\
\footnotesize{\mixtral{}} & \footnotesize{77.73\thirdcluster} & \footnotesize{76.08\thirdcluster} & \footnotesize{78.39\thirdcluster} & \footnotesize{79.57\thirdcluster} & \footnotesize{61.77\fourthcluster} & \footnotesize{76.35\thirdcluster} & \footnotesize{78.14\thirdcluster} & \footnotesize{76.06\fourthcluster} & \footnotesize{68.94\fifthcluster} \\
\cdashlinelr{1-10}
\footnotesize{\TowerInstruct{} 7B} & \footnotesize{77.61\thirdcluster} & \footnotesize{75.71\fourthcluster} & \footnotesize{78.03\thirdcluster} & \footnotesize{79.58\thirdcluster} & \footnotesize{69.25\secondcluster} & \footnotesize{\textbf{\underline{77.73}}\firstcluster} & \footnotesize{78.43\thirdcluster} & \footnotesize{77.02\secondcluster} & \footnotesize{71.53\fourthcluster} \\
\footnotesize{\TowerInstruct{} 13B} & \footnotesize{\underline{78.15}\secondcluster} & \footnotesize{\underline{76.42}\secondcluster} & \footnotesize{\underline{78.96}\secondcluster} & \footnotesize{\textbf{\underline{80.39}}\firstcluster} & \footnotesize{\textbf{\underline{70.53}}\firstcluster} & \footnotesize{\textbf{\underline{77.93}}\firstcluster} & \footnotesize{\underline{78.78}\secondcluster} & \footnotesize{\textbf{\underline{77.97}}\firstcluster} & \footnotesize{\underline{72.85}\thirdcluster} \\
\bottomrule
\\
\toprule
 & \multicolumn{9}{c}{\flores{} (xx$\rightarrow$en)} \\
Models & de & es & fr & it & ko & nl & pt & ru & zh \\
\midrule
\multicolumn{10}{l}{\small \bf Closed} \vspace{.5mm}  \\
\footnotesize{\gptthreefive{}} & \footnotesize{80.38\secondcluster} & \footnotesize{77.27\thirdcluster} & \footnotesize{80.55\thirdcluster} & \footnotesize{77.91\thirdcluster} & \footnotesize{75.22\thirdcluster} & \footnotesize{77.02\secondcluster} & \footnotesize{80.86\thirdcluster} & \footnotesize{77.73\thirdcluster} & \footnotesize{76.12\secondcluster} \\
\footnotesize{\gptfour{}} & \footnotesize{\textbf{80.74}\firstcluster} & \footnotesize{77.61\secondcluster} & \footnotesize{80.72\secondcluster} & \footnotesize{78.14\secondcluster} & \footnotesize{\textbf{76.51}\firstcluster} & \footnotesize{\textbf{77.23}\firstcluster} & \footnotesize{81.11\secondcluster} & \footnotesize{78.02\secondcluster} & \footnotesize{\textbf{76.54}\firstcluster} \\
\midrule
\multicolumn{10}{l}{\small \bf Open} \vspace{.5mm}  \\
\footnotesize{\nllb{} 54B} & \footnotesize{80.12\thirdcluster} & \footnotesize{77.09\thirdcluster} & \footnotesize{80.64\secondcluster} & \footnotesize{77.79\thirdcluster} & \footnotesize{75.32\secondcluster} & \footnotesize{76.99\secondcluster} & \footnotesize{80.81\thirdcluster} & \footnotesize{77.95\secondcluster} & \footnotesize{75.19\fourthcluster} \\
\footnotesize{\llama{} 70B} & \footnotesize{80.38\secondcluster} & \footnotesize{\textbf{\underline{77.65}}\firstcluster} & \footnotesize{80.79\secondcluster} & \footnotesize{78.05\secondcluster} & \footnotesize{75.58\secondcluster} & \footnotesize{76.77\thirdcluster} & \footnotesize{81.16\secondcluster} & \footnotesize{78.18\secondcluster} & \footnotesize{75.96\secondcluster} \\
\footnotesize{\mixtral{}} & \footnotesize{80.40\secondcluster} & \footnotesize{\textbf{\underline{77.79}}\firstcluster} & \footnotesize{80.75\secondcluster} & \footnotesize{\textbf{\underline{78.53}}\firstcluster} & \footnotesize{74.15\fourthcluster} & \footnotesize{76.87\secondcluster} & \footnotesize{80.85\thirdcluster} & \footnotesize{78.02\secondcluster} & \footnotesize{75.57\thirdcluster} \\
\cdashlinelr{1-10}
\footnotesize{\TowerInstruct{} 7B} & \footnotesize{80.17\thirdcluster} & \footnotesize{77.47\secondcluster} & \footnotesize{80.67\secondcluster} & \footnotesize{\textbf{\underline{78.40}}\firstcluster} & \footnotesize{75.62\secondcluster} & \footnotesize{76.96\secondcluster} & \footnotesize{81.30\secondcluster} & \footnotesize{78.10\secondcluster} & \footnotesize{75.68\thirdcluster} \\
\footnotesize{\TowerInstruct{} 13B} & \footnotesize{\textbf{\underline{80.55}}\firstcluster} & \footnotesize{\textbf{\underline{77.65}}\firstcluster} & \footnotesize{\textbf{\underline{81.03}}\firstcluster} & \footnotesize{\textbf{\underline{78.54}}\firstcluster} & \footnotesize{\textbf{\underline{76.53}}\firstcluster} & \footnotesize{\textbf{\underline{77.22}}\firstcluster} & \footnotesize{\textbf{\underline{81.51}}\firstcluster} & \footnotesize{\textbf{\underline{78.51}}\firstcluster} & \footnotesize{\textbf{\underline{76.46}}\firstcluster} \\
\bottomrule

\end{tabular}

%% file: tables/appendix_flores_chrf.tex
\setlength{\tabcolsep}{2.5pt}
\renewcommand{\arraystretch}{1}
\footnotesize
\begin{tabular}{lccccccccc}
\toprule
 & \multicolumn{9}{c}{\flores{} (en$\rightarrow$xx)} \\
Models & de & es & fr & it & ko & nl & pt & ru & zh \\
\midrule
\multicolumn{10}{l}{\small \bf Closed} \vspace{.5mm}  \\
\footnotesize{\gptthreefive{}} & \footnotesize{67.22\secondcluster} & \footnotesize{\textbf{57.39}\firstcluster} & \footnotesize{\textbf{72.79}\firstcluster} & \footnotesize{\textbf{60.67}\firstcluster} & \footnotesize{35.49\secondcluster} & \footnotesize{59.57\secondcluster} & \footnotesize{\textbf{72.96}\firstcluster} & \footnotesize{58.48\secondcluster} & \footnotesize{\textbf{39.21}\firstcluster} \\
\footnotesize{\gptfour{}} & \footnotesize{\textbf{67.89}\firstcluster} & \footnotesize{57.13\secondcluster} & \footnotesize{\textbf{72.89}\firstcluster} & \footnotesize{\textbf{60.60}\firstcluster} & \footnotesize{\textbf{37.18}\firstcluster} & \footnotesize{\textbf{59.97}\firstcluster} & \footnotesize{\textbf{72.98}\firstcluster} & \footnotesize{\textbf{59.50}\firstcluster} & \footnotesize{\textbf{39.32}\firstcluster} \\
\midrule
\multicolumn{10}{l}{\small \bf Open} \vspace{.5mm}  \\
\footnotesize{\nllb{} 54B} & \footnotesize{63.18\fifthcluster} & \footnotesize{55.30\fifthcluster} & \footnotesize{70.25\thirdcluster} & \footnotesize{58.83\thirdcluster} & \footnotesize{\textbf{\underline{36.54}}\firstcluster} & \footnotesize{56.99\fifthcluster} & \footnotesize{68.19\fourthcluster} & \footnotesize{57.28\thirdcluster} & \footnotesize{25.73\fifthcluster} \\
\footnotesize{\llama{} 70B} & \footnotesize{63.43\fifthcluster} & \footnotesize{55.39\fifthcluster} & \footnotesize{69.54\fourthcluster} & \footnotesize{58.20\thirdcluster} & \footnotesize{32.07\thirdcluster} & \footnotesize{56.53\fifthcluster} & \footnotesize{\underline{69.61}\secondcluster} & \footnotesize{56.58\fourthcluster} & \footnotesize{35.38\thirdcluster} \\
\footnotesize{\mixtral{}} & \footnotesize{64.14\fourthcluster} & \footnotesize{56.14\fourthcluster} & \footnotesize{\underline{70.91}\secondcluster} & \footnotesize{59.01\secondcluster} & \footnotesize{27.54\fourthcluster} & \footnotesize{56.22\sixthcluster} & \footnotesize{\underline{69.43}\secondcluster} & \footnotesize{56.07\fourthcluster} & \footnotesize{31.01\fourthcluster} \\
\cdashlinelr{1-10}
\footnotesize{\TowerInstruct{} 7B} & \footnotesize{63.87\fourthcluster} & \footnotesize{56.04\fourthcluster} & \footnotesize{70.23\thirdcluster} & \footnotesize{59.45\secondcluster} & \footnotesize{35.44\secondcluster} & \footnotesize{58.16\fourthcluster} & \footnotesize{68.74\fourthcluster} & \footnotesize{57.77\thirdcluster} & \footnotesize{35.78\thirdcluster} \\
\footnotesize{\TowerInstruct{} 13B} & \footnotesize{\underline{65.16}\thirdcluster} & \footnotesize{\underline{56.58}\thirdcluster} & \footnotesize{\underline{71.26}\secondcluster} & \footnotesize{\textbf{\underline{60.32}}\firstcluster} & \footnotesize{\textbf{\underline{37.10}}\firstcluster} & \footnotesize{\underline{59.04}\thirdcluster} & \footnotesize{69.06\thirdcluster} & \footnotesize{\underline{58.77}\secondcluster} & \footnotesize{\underline{37.40}\secondcluster} \\
\bottomrule
\\
\toprule
 & \multicolumn{9}{c}{\flores{} (xx$\rightarrow$en)} \\
Models & de & es & fr & it & ko & nl & pt & ru & zh \\
\midrule
\multicolumn{10}{l}{\small \bf Closed} \vspace{.5mm}  \\
\footnotesize{\gptthreefive{}} & \footnotesize{69.31\secondcluster} & \footnotesize{60.46\thirdcluster} & \footnotesize{69.54\secondcluster} & \footnotesize{62.76\thirdcluster} & \footnotesize{57.50\thirdcluster} & \footnotesize{60.75\secondcluster} & \footnotesize{72.56\thirdcluster} & \footnotesize{62.80\thirdcluster} & \footnotesize{58.07\secondcluster} \\
\footnotesize{\gptfour{}} & \footnotesize{\textbf{69.74}\firstcluster} & \footnotesize{61.09\secondcluster} & \footnotesize{\textbf{69.94}\firstcluster} & \footnotesize{62.75\thirdcluster} & \footnotesize{\textbf{59.55}\firstcluster} & \footnotesize{60.88\secondcluster} & \footnotesize{72.91\secondcluster} & \footnotesize{63.40\secondcluster} & \footnotesize{\textbf{58.87}\firstcluster} \\
\midrule
\multicolumn{10}{l}{\small \bf Open} \vspace{.5mm}  \\
\footnotesize{\nllb{} 54B} & \footnotesize{68.54\thirdcluster} & \footnotesize{60.72\secondcluster} & \footnotesize{69.70\secondcluster} & \footnotesize{62.95\thirdcluster} & \footnotesize{58.55\secondcluster} & \footnotesize{60.67\secondcluster} & \footnotesize{72.26\thirdcluster} & \footnotesize{62.66\thirdcluster} & \footnotesize{\textbf{\underline{58.83}}\firstcluster} \\
\footnotesize{\llama{} 70B} & \footnotesize{69.22\secondcluster} & \footnotesize{\textbf{\underline{61.34}}\firstcluster} & \footnotesize{\textbf{\underline{70.08}}\firstcluster} & \footnotesize{63.51\secondcluster} & \footnotesize{57.82\secondcluster} & \footnotesize{60.90\secondcluster} & \footnotesize{72.96\secondcluster} & \footnotesize{63.61\secondcluster} & \footnotesize{57.94\secondcluster} \\
\footnotesize{\mixtral{}} & \footnotesize{69.00\secondcluster} & \footnotesize{\textbf{\underline{61.29}}\firstcluster} & \footnotesize{69.32\secondcluster} & \footnotesize{63.38\secondcluster} & \footnotesize{55.56\fourthcluster} & \footnotesize{59.98\thirdcluster} & \footnotesize{72.18\fourthcluster} & \footnotesize{62.77\thirdcluster} & \footnotesize{56.97\thirdcluster} \\
\cdashlinelr{1-10}
\footnotesize{\TowerInstruct{} 7B} & \footnotesize{68.94\secondcluster} & \footnotesize{\textbf{\underline{61.39}}\firstcluster} & \footnotesize{69.56\secondcluster} & \footnotesize{63.59\secondcluster} & \footnotesize{58.48\secondcluster} & \footnotesize{60.65\secondcluster} & \footnotesize{73.00\secondcluster} & \footnotesize{63.37\secondcluster} & \footnotesize{57.79\secondcluster} \\
\footnotesize{\TowerInstruct{} 13B} & \footnotesize{\textbf{\underline{69.39}}\firstcluster} & \footnotesize{\textbf{\underline{61.50}}\firstcluster} & \footnotesize{\textbf{\underline{70.07}}\firstcluster} & \footnotesize{\textbf{\underline{64.06}}\firstcluster} & \footnotesize{\textbf{\underline{59.81}}\firstcluster} & \footnotesize{\textbf{\underline{61.40}}\firstcluster} & \footnotesize{\textbf{\underline{73.54}}\firstcluster} & \footnotesize{\textbf{\underline{64.41}}\firstcluster} & \footnotesize{\textbf{\underline{58.90}}\firstcluster} \\
\bottomrule
\end{tabular}

%% file: tables/flores_everything.tex
\renewcommand{\arraystretch}{1}
\footnotesize
\begin{tabular}{lccccccccc}
\toprule
 & \multicolumn{9}{c}{\flores{} (en$\rightarrow$xx)} \\
Models & de & es & fr & it & ko & nl & pt & ru & zh \\
\midrule
\multicolumn{9}{l}{\small \bf Closed} \\
\footnotesize{\gptthreefive{}} & 88.78 & 87.08 & 89.02 & 89.06 & 89.36 & 88.63 & 90.46 & 89.56 & 88.58 \\
\footnotesize{\gptfour{}} & 88.98 & 87.10 & 88.93 & 89.05 & 90.06 & 88.56 & 90.43 & 90.19 & 88.87 \\
\cdashlinelr{1-10}
\multicolumn{9}{l}{\small \bf Open} \\
\footnotesize{\nllb{} 54B} & 87.18 & 85.92 & 87.71 & 88.10 & 89.00 & 87.33 & 88.72 & 88.89 & 78.26 \\
\footnotesize{\llama{} 7B} & 84.03 & 84.37 & 85.18 & 85.18 & 80.20 & 84.48 & 87.01 & 85.09 & 82.50 \\
\footnotesize{\llama{} 13B} & 85.60 & 85.45 & 86.74 & 87.02 & 84.22 & 86.11 & 88.33 & 87.02 & 84.83 \\
\footnotesize{\llama{} 70B} & 87.31 & 86.41 & 87.82 & 88.22 & 88.07 & 87.47 & 89.11 & 88.65 & 87.32 \\
\footnotesize{\mistral{}} & 84.27 & 84.87 & 86.16 & 85.86 & 79.20 & 84.43 & 87.53 & 85.78 & 82.41 \\
\footnotesize{\mixtralBase{}} & 87.95 & 86.64 & 88.39 & 88.44 & 85.72 & 87.26 & 89.34 & 88.89 & 86.23 \\
\footnotesize{\mixtral{}} & 87.99 & 86.80 & 88.53 & 88.77 & 85.63 & 87.57 & 89.45 & 89.09 & 85.99 \\
\footnotesize{\qwen{} 72B} & 87.20 & 86.46 & 87.78 & 88.19 & 87.64 & 87.40 & 89.13 & 88.41 & 88.85 \\
\footnotesize{\gemma{} 7B} & 86.13 & 85.84 & 87.09 & 87.03 & 84.89 & 86.03 & 88.60 & 87.24 & 85.75 \\
\footnotesize{\almapretrained{} 7B} & 86.47 & 83.18 & 84.23 & 83.59 & 68.06 & 81.05 & 84.80 & 87.96 & 85.80 \\
\footnotesize{\almapretrained{} 13B} & 87.07 & 84.90 & 86.05 & 86.09 & 77.10 & 84.36 & 87.47 & 88.91 & 86.58 \\
\cdashlinelr{1-10}
\multicolumn{9}{l}{\small \bf \Tower} \\
\footnotesize{\TowerBase{} 7B} & 86.91 & 85.95 & 87.76 & 87.93 & 86.55 & 87.37 & 89.47 & 88.72 & 86.48 \\
\footnotesize{\TowerBase{} 13B} & 87.21 & 86.01 & 88.34 & 88.25 & 88.78 & 87.52 & 89.36 & 88.30 & 87.14 \\
\footnotesize{\TowerInstruct{} 7B} & 87.82 & 86.76 & 88.44 & 88.73 & 89.41 & 88.38 & 89.60 & 89.53 & 87.90 \\
\footnotesize{\TowerInstruct{} 13B} & 88.16 & 87.06 & 88.92 & 89.21 & 89.92 & 88.63 & 89.78 & 89.95 & 88.29 \\
\bottomrule
\\
\toprule
 & \multicolumn{9}{c}{\flores{} (xx$\rightarrow$en)} \\
Models & de & es & fr & it & ko & nl & pt & ru & zh \\
\midrule
\multicolumn{9}{l}{\small \bf Closed} \\
\footnotesize{\gptthreefive{}} & 89.60 & 87.26 & 89.46 & 88.03 & 87.83 & 87.71 & 89.78 & 86.69 & 86.92 \\
\footnotesize{\gptfour{}} & 89.76 & 87.57 & 89.61 & 88.21 & 88.58 & 87.88 & 89.94 & 86.94 & 87.29 \\
\cdashlinelr{1-10}
\multicolumn{9}{l}{\small \bf Open} \\
\footnotesize{\nllb{} 54B} & 89.17 & 87.25 & 89.29 & 87.91 & 87.86 & 87.49 & 89.38 & 86.66 & 86.55 \\
\footnotesize{\llama{} 7B} & 88.47 & 86.63 & 88.78 & 87.48 & 85.52 & 86.67 & 88.98 & 85.87 & 85.53 \\
\footnotesize{\llama{} 13B} & 89.01 & 86.98 & 89.14 & 87.87 & 86.95 & 87.23 & 89.26 & 86.37 & 86.35 \\
\footnotesize{\llama{} 70B} & 89.44 & 87.49 & 89.55 & 88.18 & 87.91 & 87.52 & 89.84 & 86.87 & 86.91 \\
\footnotesize{\mistral{}} & 88.83 & 87.07 & 88.81 & 87.69 & 85.16 & 86.93 & 89.05 & 86.21 & 85.65 \\
\footnotesize{\mixtralBase{}} & 89.55 & 87.57 & 89.58 & 88.35 & 87.03 & 87.54 & 89.80 & 86.79 & 86.63 \\
\footnotesize{\mixtral{}} & 89.57 & 87.65 & 89.56 & 88.44 & 87.37 & 87.54 & 89.73 & 86.81 & 86.88 \\
\footnotesize{\qwen{} 72B} & 89.67 & 87.66 & 89.58 & 88.41 & 88.42 & 87.72 & 89.88 & 87.13 & 87.94 \\
\footnotesize{\gemma{} 7B} & 89.17 & 87.09 & 89.12 & 87.81 & 87.28 & 87.23 & 89.48 & 86.59 & 86.59 \\
\footnotesize{\almapretrained{} 7B} & 89.23 & 86.84 & 89.01 & 87.68 & 83.35 & 86.92 & 89.05 & 86.81 & 86.59 \\
\footnotesize{\almapretrained{} 13B} & 89.81 & 87.42 & 89.42 & 88.18 & 86.26 & 87.59 & 89.70 & 87.23 & 87.16 \\
\cdashlinelr{1-10}
\multicolumn{9}{l}{\small \bf \Tower} \\
\footnotesize{\TowerBase{} 7B} & 89.26 & 87.15 & 89.47 & 88.14 & 87.80 & 87.45 & 89.77 & 86.41 & 86.72 \\
\footnotesize{\TowerBase{} 13B} & 89.54 & 87.42 & 89.55 & 88.11 & 88.24 & 87.61 & 89.71 & 86.18 & 87.02 \\
\footnotesize{\TowerInstruct{} 7B} & 89.48 & 87.48 & 89.50 & 88.39 & 88.16 & 87.66 & 89.92 & 86.90 & 86.96 \\
\footnotesize{\TowerInstruct{} 13B} & 89.61 & 87.62 & 89.67 & 88.42 & 88.48 & 87.92 & 90.07 & 87.20 & 87.27 \\
\bottomrule
\end{tabular}

%% file: tables/wmt_everything.tex
\renewcommand{\arraystretch}{1}
\footnotesize
\begin{tabular}{lccccccccc}
\toprule
 & \multicolumn{6}{c}{\wmt{}}\\
Models & en$\rightarrow$de & en$\rightarrow$ru & en$\rightarrow$zh & de$\rightarrow$en & ru$\rightarrow$en & zh$\rightarrow$en \\
\midrule
\multicolumn{6}{l}{\small \bf Closed} \\
\footnotesize{\gptthreefive{}} & 84.61 & 85.38 & 86.70 & 85.91 & 83.02 & 81.52 \\
\footnotesize{\gptfour{}} & 84.89 & 86.07 & 87.08 & 86.17 & 83.63 & 81.27 \\
\cdashlinelr{1-7}
\multicolumn{6}{l}{\small \bf Open} \\
\footnotesize{\nllb{} 54B} & 77.40 & 83.91 & 74.48 & 80.06 & 80.52 & 76.60 \\
\footnotesize{\llama{} 7B} & 75.02 & 77.87 & 79.16 & 83.36 & 80.58 & 77.40 \\
\footnotesize{\llama{} 13B} & 78.29 & 80.44 & 81.30 & 83.92 & 81.54 & 78.73 \\
\footnotesize{\llama{} 70B} & 81.62 & 83.04 & 84.19 & 85.12 & 82.84 & 79.73 \\
\footnotesize{\mistral{}} & 76.78 & 80.27 & 81.26 & 84.18 & 81.52 & 79.11 \\
\footnotesize{\mixtralBase{}} & 81.92 & 83.39 & 83.81 & 85.04 & 82.70 & 79.50 \\
\footnotesize{\mixtral{}} & 83.07 & 83.79 & 83.94 & 85.45 & 83.02 & 80.04 \\
\footnotesize{\qwen{} 72B} & 81.44 & 83.31 & 86.48 & 85.54 & 83.01 & 80.60 \\
\footnotesize{\gemma{} 7B} & 79.56 & 82.20 & 83.56 & 84.60 & 82.14 & 79.24 \\
\footnotesize{\almapretrained{} 7B} & 80.20 & 83.01 & 82.68 & 83.51 & 81.82 & 78.66 \\
\footnotesize{\almapretrained{} 13B} & 81.18 & 83.72 & 83.83 & 84.32 & 82.71 & 79.22 \\
\footnotesize{\almar{} 7B} & 82.41 & 84.28 & 83.51 & 84.55 & 82.50 & 80.13 \\
\footnotesize{\almar{} 13B} & 83.59 & 85.37 & 84.43 & 85.39 & 83.23 & 80.48 \\
\cdashlinelr{1-7}
\multicolumn{6}{l}{\small \bf \Tower} \\
\footnotesize{\TowerBase{} 7B} & 81.03 & 83.25 & 84.00 & 84.09 & 80.08 & 78.92 \\
\footnotesize{\TowerBase{} 13B} & 81.18 & 83.46 & 84.03 & 83.89 & 80.03 & 78.94 \\
\footnotesize{\TowerInstruct{} 7B} & 83.22 & 84.73 & 84.89 & 85.24 & 82.94 & 80.13 \\
\footnotesize{\TowerInstruct{} 13B} & 83.98 & 85.51 & 85.92 & 85.62 & 83.21 & 80.72 \\
\bottomrule
\end{tabular}

%% file: tables/tico_everything.tex
\renewcommand{\arraystretch}{1}
\footnotesize
\begin{tabular}{lccccccccc}
\toprule
 & \multicolumn{5}{c}{\tico{}}\\
Models & en$\rightarrow$es & en$\rightarrow$fr & en$\rightarrow$pt & en$\rightarrow$ru & en$\rightarrow$zh \\
\midrule
\multicolumn{5}{l}{\small \bf Closed} \\
\footnotesize{\gptthreefive{}} & 88.67 & 81.86 & 90.30 & 87.88 & 88.09 \\
\footnotesize{\gptfour{}} & 88.76 & 81.85 & 90.30 & 88.36 & 88.32 \\
\cdashlinelr{1-6}
\multicolumn{5}{l}{\small \bf Open} \\
\footnotesize{\nllb{} 54B} & 88.74 & 82.01 & 89.84 & 88.67 & 85.97 \\
\footnotesize{\llama{} 7B} & 85.77 & 78.08 & 86.97 & 82.99 & 81.86 \\
\footnotesize{\llama{} 13B} & 86.94 & 79.83 & 88.48 & 85.44 & 84.89 \\
\footnotesize{\llama{} 70B} & 87.84 & 80.67 & 89.24 & 87.12 & 87.44 \\
\footnotesize{\mistral{}} & 86.25 & 79.18 & 87.87 & 84.35 & 84.13 \\
\footnotesize{\mixtralBase{}} & 88.12 & 81.15 & 89.27 & 87.14 & 86.58 \\
\footnotesize{\mixtral{}} & 88.23 & 81.39 & 89.48 & 87.04 & 86.84 \\
\footnotesize{\qwen{} 72B} & 86.08 & 80.32 & 88.20 & 80.53 & 86.68 \\
\footnotesize{\gemma{} 7B} & 87.30 & 78.20 & 88.66 & 86.16 & 86.78 \\
\footnotesize{\almapretrained{} 7B} & 84.42 & 76.74 & 84.92 & 86.53 & 85.27 \\
\footnotesize{\almapretrained{} 13B} & 86.17 & 79.09 & 87.56 & 87.27 & 86.54 \\
\footnotesize{\almar{} 7B} & 84.63 & 76.02 & 82.92 & 87.80 & 85.41 \\
\footnotesize{\almar{} 13B} & 85.93 & 79.90 & 87.41 & 88.58 & 86.22 \\
\cdashlinelr{1-6}
\multicolumn{5}{l}{\small \bf \Tower} \\
\footnotesize{\TowerBase{} 7B} & 87.90 & 81.20 & 89.45 & 86.94 & 86.97 \\
\footnotesize{\TowerBase{} 13B} & 87.90 & 81.48 & 89.54 & 87.26 & 87.57 \\
\footnotesize{\TowerInstruct{} 7B} & 88.34 & 81.60 & 89.38 & 88.11 & 87.63 \\
\footnotesize{\TowerInstruct{} 13B} & 88.63 & 81.82 & 89.48 & 88.49 & 88.20 \\
\bottomrule
\end{tabular}

%% file: tables/appendix_ape.tex
\renewcommand{\arraystretch}{1}
\footnotesize
\begin{tabular}{lcc}
\toprule
 & \multicolumn{2}{c}{APE }\\
Models & en$\rightarrow$xx & xx$\rightarrow$en \\
\midrule
\footnotesize{Baseline (no edits)} & \footnotesize{78.84\fourthcluster} & \footnotesize{78.80\fourthcluster} \\
\cdashlinelr{1-3}
\footnotesize{\gptthreefive{}} & \footnotesize{82.32\thirdcluster} & \footnotesize{77.91\fifthcluster} \\
\footnotesize{\gptfour{}} & \footnotesize{\textbf{85.52}\firstcluster} & \footnotesize{\textbf{83.12}\firstcluster} \\
\midrule
\footnotesize{\TowerInstruct{} 7B} & \footnotesize{83.10\thirdcluster} & \footnotesize{80.19\thirdcluster} \\
\footnotesize{\TowerInstruct{} 13B} & \footnotesize{\underline{83.65}\secondcluster} & \footnotesize{\underline{80.89}\secondcluster} \\
\bottomrule
\end{tabular}

%% file: tables/appendix_gec.tex
\renewcommand{\arraystretch}{1}
\footnotesize
\begin{tabular}{lc}
\toprule
 & GEC\\
Models & Multilingual \\
\midrule
\multicolumn{2}{l}{\small \bf Closed} \vspace{.5mm}  \\
\footnotesize{\gptthreefive{}} & \footnotesize{\textbf{0.49}\firstcluster} \\
\footnotesize{\gptfour{}} & \footnotesize{0.48\thirdcluster} \\
\midrule
\multicolumn{2}{l}{\small \bf Open} \vspace{.5mm}  \\
\footnotesize{\llama{} 70B} & \footnotesize{\underline{0.43}\fourthcluster} \\
\footnotesize{\mixtral{}} & \footnotesize{\underline{0.43}\fourthcluster} \\
\cdashlinelr{1-2}
\footnotesize{\TowerInstruct{} 7B} & \footnotesize{\underline{0.42}\fourthcluster} \\
\footnotesize{\TowerInstruct{} 13B} & \footnotesize{\underline{0.43}\fourthcluster} \\
\bottomrule
\end{tabular}

%% file: tables/appendix_ape_everything.tex
\renewcommand{\arraystretch}{1}
\footnotesize
\begin{tabular}{lccccccccc}
\toprule
 & \multicolumn{5}{c}{\wmt{}}\\
Models & en$\rightarrow$de & en$\rightarrow$ru & en$\rightarrow$zh & de$\rightarrow$en & ru$\rightarrow$en & zh$\rightarrow$en \\
\midrule
\footnotesize{Baseline (no edits)} & 77.87 & 82.93 & 75.72 & 79.92 & 80.05 & 76.44 \\
\cdashlinelr{1-7}
\multicolumn{6}{l}{\small \bf Closed} \\
\footnotesize{\gptthreefive{}} & 80.67 & 84.03 & 82.27 & 78.48 & 78.88 & 76.37 \\
\footnotesize{\gptfour{}} & 84.65 & 86.15 & 85.75 & 85.39 & 83.21 & 80.75 \\
\cdashlinelr{1-7}
\multicolumn{6}{l}{\small \bf Open} \\
\footnotesize{\gptthreefive{}} & 80.67 & 84.03 & 82.27 & 78.48 & 78.88 & 76.37 \\
\footnotesize{\gptfour{}} & 84.65 & 86.15 & 85.75 & 85.39 & 83.21 & 80.75 \\
\footnotesize{\llama{} 70B} & 78.49 & \multicolumn{1}{c}{---} & 78.20 & 81.30 & 80.76 & \multicolumn{1}{c}{---} \\
\footnotesize{\mixtral{}} & 82.12 & \multicolumn{1}{c}{---} & 83.15 & 83.40 & 82.22 & \multicolumn{1}{c}{---} \\
\cdashlinelr{1-7}
\multicolumn{6}{l}{\small \bf \Tower} \\
\footnotesize{\TowerInstruct{} 7B} & 81.86 & 83.92 & 83.52 & 82.29 & 80.82 & 77.45 \\
\footnotesize{\TowerInstruct{} 13B} & 82.03 & 84.34 & 84.59 & 83.22 & 81.30 & 78.15 \\
\bottomrule
\end{tabular}

%% file: tables/appendix_gec_everything.tex
\renewcommand{\arraystretch}{1}
\footnotesize
\begin{tabular}{lccc}
\toprule
Models & en & de & es \\
\midrule
\footnotesize{Baseline (no edits)} & 13.75 & 18.23 & 18.00 \\
\cdashlinelr{1-4}
\multicolumn{4}{l}{\small \bf Closed} \\
\footnotesize{\gptthreefive{}} & 14.71 & 13.19 & 17.29 \\
\footnotesize{\gptfour{}} & 16.48 & 12.89 & 15.86 \\
\cdashlinelr{1-4}
\multicolumn{4}{l}{\small \bf Open} \\
\footnotesize{\llama{} 70B} & 17.46 & 20.67 & 27.09 \\
\footnotesize{\mixtral{}} & 16.44 & 15.38 & 19.47 \\
\cdashlinelr{1-4}
\multicolumn{4}{l}{\small \bf \Tower} \\
\footnotesize{\TowerInstruct{} 7B} & 13.39 & 14.77 & 17.23 \\
\footnotesize{\TowerInstruct{} 13B} & 13.13 & 14.42 & 19.48 \\
\bottomrule
\end{tabular}

%% file: tables/appendix_ner_everything.tex
\renewcommand{\arraystretch}{1}
\footnotesize
\begin{tabular}{lccccccc}
\toprule
Models & en & de & es & fr & it & pt & zh \\
\midrule
\multicolumn{8}{l}{\small \bf Closed} \\
\footnotesize{\gptthreefive{}} & 55.43 & 60.12 & 56.82 & 53.34 & 55.46 & 52.57 & 17.82 \\
\footnotesize{\gptfour{}} & 63.61 & 66.58 & 65.24 & 58.72 & 63.39 & 61.74 & 39.88 \\
\cdashlinelr{1-8}
\multicolumn{8}{l}{\small \bf Open} \\
\footnotesize{\llama{} 70B} & 46.34 & 48.79 & 50.69 & 47.50 & 53.96 & 45.60 & 19.44 \\
\footnotesize{\mixtral{}} & 45.74 & 46.94 & 46.03 & 46.11 & 50.86 & 40.21 & 16.51 \\
\cdashlinelr{1-8}
\multicolumn{8}{l}{\small \bf \Tower} \\
\footnotesize{\TowerInstruct{} 7B} & 75.09 & 78.01 & 74.89 & 70.35 & 76.39 & 73.88 & 53.13 \\
\footnotesize{\TowerInstruct{} 13B} & 77.52 & 79.73 & 76.69 & 74.55 & 80.36 & 77.47 & 56.57 \\
\bottomrule
\end{tabular}